\documentclass[pdflatex,sn-nature]{sn-jnl}

\usepackage{graphicx}%
\usepackage{multirow}%
\usepackage{amsmath,amssymb,amsfonts}%
\usepackage{amsthm}%
\usepackage{mathrsfs}%
\usepackage[title]{appendix}%
\usepackage{xcolor}%
\usepackage{textcomp}%
\usepackage{manyfoot}%
\usepackage{booktabs}%
\usepackage{algorithm}%
\usepackage{algorithmicx}%
\usepackage{algpseudocode}%
\usepackage{listings}%
\usepackage{mathtools}%

\usepackage{tikz}
\usetikzlibrary{arrows.meta}
\usetikzlibrary{calc}

\theoremstyle{thmstyleone}%
\theoremstyle{thmstyletwo}%

\theoremstyle{thmstylethree}%

\newcommand{\Standard}{\textsc{Standard}}
\newcommand{\OGReorder}{\textsc{Original-Reorder}}
\newcommand{\EReorder}{\textsc{EquiDist-Reorder}}
\newcommand{\UDReorder}{\textsc{Uniform-Reorder}}
\newcommand{\NPBReorder}{\textsc{NegBias-Reorder}}
\newcommand{\LSDReorder}{\textsc{LeftSkew-Reorder}}

\algnewcommand\algorithmicforeach{\textbf{for each}}
\algdef{S}[FOR]{ForEach}[1]{\algorithmicforeach\ #1\ \algorithmicdo}

\begin{document}

\title[Analysing the Influence of Reorder Strategies for Cartesian Genetic Programming]{Analysing the Influence of Reorder Strategies for Cartesian Genetic Programming}


\author*[1]{\fnm{Henning} \sur{Cui}}\email{henning.cui@uni-a.de}

\author[2]{\fnm{Andreas} \sur{Margraf}}\email{andreas.margraf@igcv.fraunhofer.de}

\author[1]{\fnm{J\"org} \sur{H\"ahner}}\email{joerg.haehner@uni-a.de}

\affil*[1]{\orgdiv{Organic Computing Group}, \orgname{Institute for Computer Science, University of Augsburg}, \orgaddress{\street{Am Technologiezentrum 8}, \city{Augsburg}, \postcode{86159}, \state{Bavaria}, \country{Germany}}}

\affil[2]{\orgname{Fraunhofer IGCV}, \orgaddress{\street{Am Technologiezentrum 2}, \city{Augsburg}, \postcode{86159}, \state{Bavaria}, \country{Germany}}}

\abstract{
    \emph{Cartesian Genetic Programming} (CGP) suffers from a specific limitation: The \emph{positional bias}, a phenomenon in which mostly genes at the start of the genome contribute to a program output, while genes at the end rarely do.
    This can lead to an overall worse performance of CGP.
    One solution to overcome the positional bias is to introduce \emph{reordering methods}, which shuffle the current genotype without changing its corresponding phenotype.
    There are currently two different reorder operators that extend the classic CGP formula and improve its fitness value.
    In this work, we discuss possible shortcomings of these two existing operators.
    Afterwards, we introduce \emph{three novel operators} which reorder the genotype of a graph defined by CGP.
    We show empirically on four Boolean and four symbolic regression benchmarks that the number of iterations until a solution is found and/or the fitness value improves by using CGP with a reorder method.
    However, there is no consistently best performing reorder operator.
    Furthermore, their behaviour is analysed by investigating their convergence plots and we show that all behave the same in terms of convergence type.
    }




\maketitle

\section{Introduction}
\label{sec:introduction}

\emph{Cartesian Genetic Programming} (CGP) is a form of \emph{Genetic Programming} (GP) and a nature inspired \emph{search heuristic}. 
The standard CGP version is represented by a directed, acyclic and feed-forward graph---instead of a tree structured representation, as is the case in GP.
CGP was introduced by Miller~\cite{cgp_origin} to evolve digital circuits, which is still an active research field as of today~\cite{Froehlich22,Manazir22}.
Furthermore, its concept is used in a diverse field of problem domains like classification or regression~\cite{cgp_book}.
Another more prominent application for CGP is on the subject of neural architecture search~\cite{Torabi2022,Suganuma2020}.

While CGP has some advantages over GP---for example the absence of bloat~\cite{cgp_origin}---it has its own specific problems.
One issue is the \emph{positional bias}~\cite{dag_origin}, which describes a limitation of CGP to fully explore its search space.
A possible solution to this problem is called \emph{Reorder}, an operator which shuffles CGP's genome ordering without changing its phenotype~\cite{dag_origin,dag_extension}.
However, Cui et al.~\cite{cui_ecta_reorder} showed that the original reorder method does not fully mitigate the positional bias but only lessens it.
They changed the original operator so that nodes that should be reordered are spaced \emph{equidistantly apart}.
This fully eliminated the positional bias and lead to an improvement in fitness.
However, upon further inspection, enforcing an equidistant spacing brings the possibility that other biases and problems arise.
Hence, in this work we will discuss possible issues as well as evaluate and introduce other novel reordering methods.
We show empirically that CGP always profits from some reorder method, as it will lead to a better fitness value and/or decrease the number of iterations until a solution is found.
Furthermore, by presenting and analysing convergence plots, we show that the behaviour of CGP will not change by introducing some form of reordering.

In this work, we further expand on the \emph{Equidistant Reorder} method, which spaces specific nodes equidistantly apart. 
We extend our previous work on this topic originally presented at ECTA/IJCCI 2023~\cite{cui_ecta_reorder}.
Based on the previous work, we introduce new reordering methods and provide a broader empirical analysis by extending our set of benchmarks to include symbolic regression.
Furthermore, the behaviour of all CGP algorithms used is inspected via convergence plots.

We provide a quick overview of the core principles of CGP in Section~\ref{sec:cgp}.
Afterwards, Section~\ref{sec:related_work} gives an overview of related work.
We then reintroduce existing reorder operators and present our novel operators in Section~\ref{sec:reorder_strategies}.
In the following Section~\ref{sec:evaluation}, their performance and behaviour is evaluated.
At last, our findings and discussions for future research directions are summarized in Section~\ref{sec:conclusion}.

\section{Cartesian Genetic Programming}
\label{sec:cgp}

This section reintroduces the supervised learning algorithm called Cartesian Genetic Programming.
An additional focus point will be the \emph{positional bias} which impacts the search behaviour of CGP negatively.

\subsection{Representation}
CGP is represented by a directed, acyclic and feed-forward graph.
It contains \emph{nodes}, which are arranged in a $c \times r$ grid, with $c \in \mathbb{N}_+$ and $r \in \mathbb{N}_+$ defining the number of columns and rows in the grid respectively~\cite{cgp_origin}.
Using nodes, it is possible to feed-forward an arbitrary amount of program inputs via partially connected nodes to calculate any desired amount of outputs.
However, with today's standard, a CGP model consists of only one row for most applications~\cite{cgp_book}.
That means, $r=1$.

The set of nodes in a graph defined by CGP can be divided into input-, output- and computational nodes.
Input nodes directly receive the program input and relay these values to computational and/or output nodes.
Differently, computational nodes are \emph{represented} by three genes: one \emph{function gene} and $m$ connection genes, with $m \in \mathbb{N}_+$ being the maximum \emph{arity} of one function in the whole function set.
Such connection genes define a path between a previous and the current node, and a function genes addresses the encoded computational function of a node.
If a function needs less than $m$ inputs, all excess input genes are ignored.
Contrary, output nodes define the programs final output and are \emph{represented} by a single connection gene in the genotype.
They redirect the output of an input- or computational node for that reason.

Another important distinction for this work is the categorization of computation nodes into two groups: \emph{active} and \emph{inactive} nodes. 
The former are nodes which contribute to the program's final output because they are part of a path to one or multiple output nodes.
The latter are not part of a path to the output nodes, hence they do not contribute to an output.
However, by allowing such inactive nodes to persist throughout the training process, it may improve CGP?s evolutionary search through neutral genetic drift~\cite{Miller2006,neutral_drift}. 

Given this description of CGP's representation, when we mention a graph with $N \in \mathbb{N}_+$ nodes, this graph will have only one row, $N$ computational nodes, as well as additional input and output nodes corresponding to the given learning task.
Furthermore, to improve readability and clarity, a graph defined by this standard representation will be called \Standard{}.

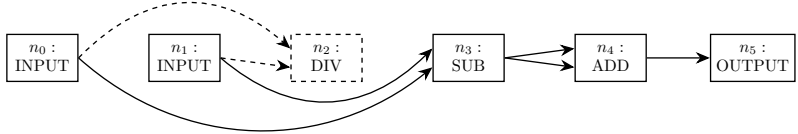
\begin{figure}[t]
    \centering
    \resizebox{0.8\linewidth}{!}{%
        \begin{tikzpicture}
            [>={Stealth[width=2.5mm,length=3mm]},
            thick,
            cgpnode/.style={rectangle, draw, minimum width = 1.5cm, minimum height = 1cm, align=center}]
            \node (input1) at (0,0) [cgpnode] {$n_0:$ \\ INPUT};
            \node (input2) at (3,0) [cgpnode] {$n_1:$ \\ INPUT};
            \node (add) at (6,0) [cgpnode, dashed] {$n_2:$ \\ DIV};
            \node (sub) at (9,0) [cgpnode] {$n_3:$ \\ SUB};
            \node (add2) at (12,0) [cgpnode] {$n_4:$ \\ ADD};
            \node (out) at (15,0) [cgpnode] {$n_5:$ \\ OUTPUT};
            
            \draw[->, dashed] (input1.east) .. controls ($(input2.north west) + (0, 1)$) and ($(input2.north east) + (0, 1)$) .. ($(add.west) + (0, 0.2)$);
            \draw[->] (input1.east) .. controls ($(input2.south) - (0, 1.5)$) and ($(add.south) - (0, 1.5)$) .. ($(sub.west) - (0, 0.2)$);
            \draw[->, dashed] (input2.east) to ($(add.west) - (0, 0.2)$);
            \draw[->] (input2.east) .. controls ($(add.south west) - (0, 0.75)$) and ($(add.south east) - (0, 0.75)$) .. ($(sub.west) + (0, 0.2)$);
            \draw[->] (sub.east) to ($(add2.west) + (0, 0.2)$);
            \draw[->] (sub.east) to ($(add2.west) - (0, 0.2)$);
            \draw[->] (add2) to (out);
        \end{tikzpicture}
    }
    \caption{
        Example graph defined by a CGP genotype.
        The dashed node and connections are \emph{inactive} due to not contributing to the output.
    }
    \label{fig:cgp_example}
\end{figure}

In Figure~\ref{fig:cgp_example}, an illustrative example of a graph with $c = 6$, $r=1$ and $N=3$ defined by a CGP genotype is depicted.
Active nodes are drawn with a solid line, while inactive nodes are marked by dashed lines.
The first two nodes ($n_0$ and $n_1$) are input nodes and correspond to two different program inputs. 
The following nodes ($n_2$--$n_4$) are computational nodes, and one output node ($n_5$) defines the program's output. 
In this example, both inputs are subtracted.
Afterwards, this intermediate result is being added to itself and redirected as the program output. 
The node $n_2$ does not contains a path to an output node and is inactive.

\subsection{Positional Bias}
\label{subsec:positional_bias}
Enforcing a feed-forward grid leads to one major negative impact of CGP's evolutionary search: The \emph{positional bias}, which was found by Goldman and Punch~\cite{dag_origin}.
Their findings show that the probability of a computational node being active is not uniformly distributed throughout the whole graph. 
Instead, computational nodes closer to the input nodes have a \emph{higher chance} of being active compared to computational nodes near the output nodes.
With this negative impact, the positional bias makes it difficult to solve certain tasks and could increase the number of iterations until a solution is found~\cite{dag_origin,struct_bias}.

\begin{figure}[t]%
    \centering
    \includegraphics[width=0.8\textwidth]{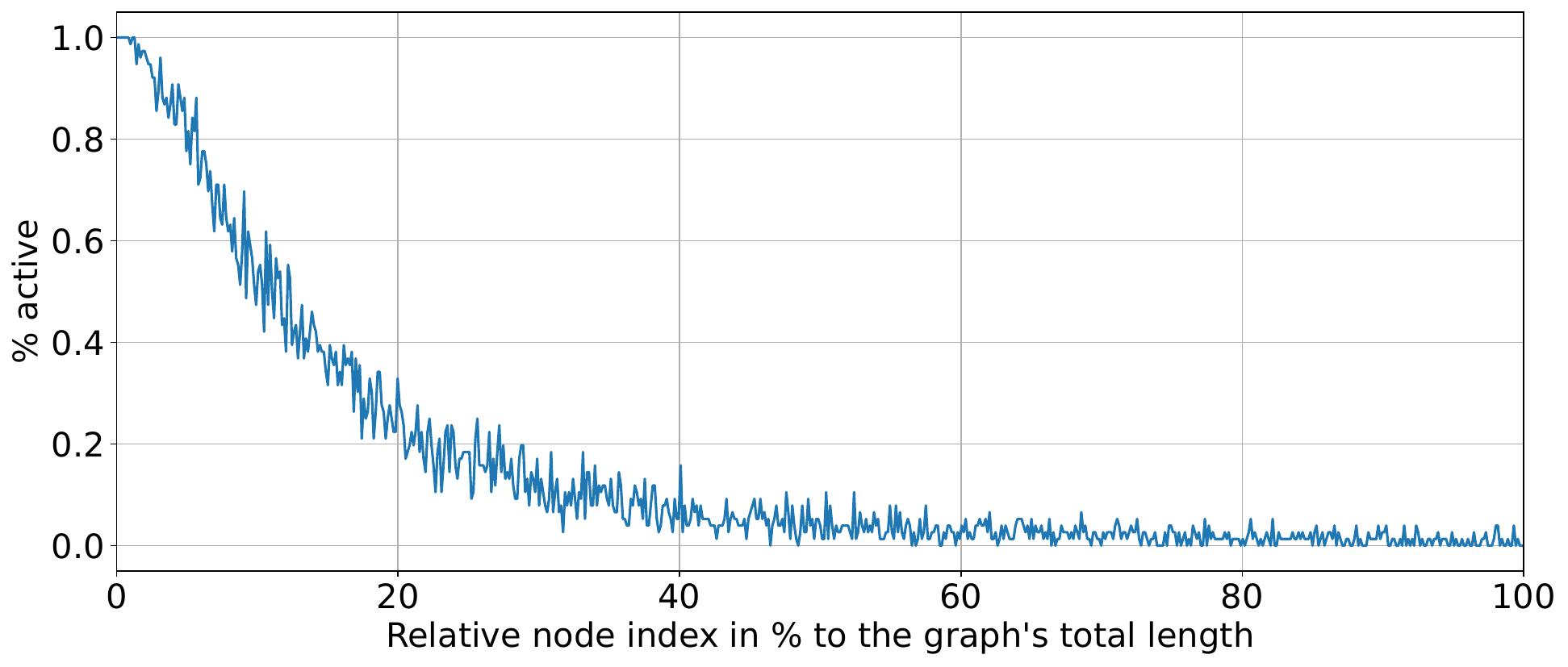}
    \caption{Distribution of active nodes over a graph defined by CGP. Example generated over 75 independent runs on the 3-bit multiply Boolean benchmark.}
    \label{fig:ex_pos_bias}
\end{figure}

This bias can be visualized for a better understanding, as is shown in Figure~\ref{fig:ex_pos_bias}.
The graph shows the distribution of active nodes, averaged over 75 independent runs on the \emph{3-bit multiply} Boolean benchmark.
On the x-axis, the position of a computational node in its graph is given.
The y-axis indicates its probability of being active.
Here, the effects of the positional bias can be seen.
About the first fifth of computational nodes have a (very) high probability of being active.
However, the probability of being active for the remaining nodes are minimal.
As a consequence, their mutations and usefulness cannot be evaluated regularly.

\section{Related Work}
\label{sec:related_work}

In our work, we focus on improving the \emph{Reorder} operator from Goldman and Punch~\cite{dag_origin}.
The motivation behind their work was mitigating the aforementioned positional bias~\cite{dag_origin,dag_extension}.
To the best of our knowledge, only Cui et al.~\cite{cui_ecta_reorder} extended the concept of Reorder.
In their work, the authors reordered active nodes by placing them equidistantly apart.
This lead to an improved fitness value.

Additionally, various other articles layed out the foundation for this work.
The most important ones for this work introduced novel operators for CGP and/or focused on various other limitations in CGP's ability to explore the search space.

Instead of reordering active nodes, Kalkreuth~\cite{phenotypic_duplication} swapped genes in the phenotype.
For that reason, two new operators were introduced which were able to find solutions in less iterations.

Walker and Miller~\cite{modular_cgp} also explicitly changed CGP's geno- and phenotype.
With their extension, nodes or subgraphs in the genotype can be dynamically created, evolved, reused and removed.
They were able to improve CGP's fitness value, with the addition of less training iterations needed.

It is also possible to extend the function pool of CGP, as was done by Harding et al.~\cite{self_modyfing}.
They included functions which directly change CGP's phenotype.
This allowed them to solve problems with extremely large numbers of inputs.

The representation of a graph defined by CGP can be converted into an integer representation.
Wilson et al.~\cite{positional_cgp}, however, introduced a floating point representation.
This allowed them to use specialized operators to change the genotype.
In this way, they were able to add or remove whole subgraphs.

Another method to mitigate the positional bias is to allow arbitrary connections as long as no cycles form~\cite{dag_origin,dag_extension}.
However, Cui et al.~\cite{Cui2023} found that this may lead to other problems and, consequently, the search space is not explored optimally.
To solve this problem, they introduced weights to guide the mutation of connections, which can reduce the positional bias for \Standard{}.

A different route is to decrease the size of the search space, as was done by Suganuma et al.~\cite{Suganuma2020}.
They used a highly specific function set for neural architecture search for that goal.
With another method, Torabi et al.~\cite{Torabi2022} used a specific crossover operator to decrease the size of the search space which uses specific domain knowledge.

\section{Reorder Strategies}
\label{sec:reorder_strategies}

As previously mentioned in Section~\ref{subsec:positional_bias}, enforcing a feed-forward grid leads to the positional bias.
To mitigate these negative effects, Goldman and Punch~\cite{dag_origin} introduced two new extensions to CGP.
In this work, we build upon one of those two: \emph{Reorder}, an extension which shuffles the genotype without changing the phenotype.
For that reason, the original Reorder strategy~\cite{dag_origin} is reintroduced for better understanding.
We also introduce several novel reorder strategies, which may completely remove the positional bias or deliberately introduce other skewed distributions of active nodes.

Please note that all operators change CGP's genotype.
However, by respecting the sequence of operations of active nodes, the phenotype---and thereby the program's output---stays the same. 

\subsection{Original Reorder}
\label{subsec:reorder_og}
The original reorder operator is able to mitigate the positional bias~\cite{dag_origin,dag_extension}.
However, we argue that it still suffers from a lessened positional bias.

\subsubsection{Reorder Strategy}
The original Reorder operator begins by preparing a dependency set $D~\coloneqq~\left\{ \left(a, b\right) \mid a\text{ is input or computational node, } b\text{ is computational node} \right\}$.
It contains information about each computational node and the nodes from which it gets its input from.
Furthermore, a new genome $\mathcal{G}^\prime$ must be initialized, containing all input and output nodes from the original genome.
As input and output nodes should not be assigned a new position, they will not be modified in the shuffled genotype.

Now, in order to be able to proceed with the shuffling of nodes, the concept of \emph{satisfying a nodes dependencies} must be introduced.
Let $m$ be the arity, $b$ be a computational node, and $a_i$ for $i=1, \cdots, m$ be an arbitrary node from which $b$ gets its input from, that means $(a_i, b) \in D$.
This indicates a directed edge connecting $a_i$ to $b$ via the connection gene of $b$ for $i=1, \cdots, m$.
Furthermore, let $A_b \coloneqq \left\{a_1, \cdots, a_m\right\}$ be the set of all nodes from which $b$ gets its input from.
Given this information, $b$ is \emph{satisfied} when all nodes in $A\textbf{}$ are mapped into the new genome $\mathcal{G}^\prime$.

With the introduction of satisfied nodes, the set of \emph{addable} nodes $Q$ can be created.
This set is initialized by adding all computational nodes whose dependencies are \emph{already satisfied} in $\mathcal{G}^\prime$, that means: $Q \coloneqq \left\{b \mid b \text{ is computational node in  } \mathcal{G}, A_b \subseteq G^\prime\right\}$.

Given both sets $D$ and $Q$, a genotype can be reordered by repeating the following steps:

\begin{enumerate}
    \item Select and remove a random node $n \in Q$.
    \item Map $n$ sequentially into the next free location in $\mathcal{G}^\prime$.
    \item For each node pair $(n, b) \in D$ with node $b$ depending on $n$, do the following: If all dependencies of $b$ are satisfied in $\mathcal{G}^\prime$, add $b$ to $Q$.
    \item If all computational nodes have been assigned a new position in $\mathcal{G}^\prime$, stop. Else, go to Step 1.
\end{enumerate}

To better illustrate this operator, its workings are also depicted in Algorithm~\ref{algo:reorder_og}. 
Please note: When all computational nodes have been shuffled into $\mathcal{G}^\prime$, $Q$ must be an empty set, too.
Furthermore, to improve readability in the next sections, the original reorder operator will be called \OGReorder{}.

\begin{algorithm}[t]
    \caption{Original Reorder}\label{algo:reorder_og}
    \begin{algorithmic}[1]
        \State Initialize dependency set $D$
        \State Initialize new genome $\mathcal{G}^\prime$
        \State Initialize addable set $Q$
        \While{$Q \neq \emptyset$}
        \State Select and remove a random node $n \in Q$
        \State Insert $n$ into the next free location in $\mathcal{G}^\prime$
        \State Update $Q$
        \EndWhile
    \end{algorithmic}
\end{algorithm}

\subsubsection{Reorders Limitation}
In an earlier work, Cui et al.~\cite{cui_ecta_reorder} showed empirically that \OGReorder{} does not fully eliminate the positional bias but only lessens it.
This is due to the constraint of a nodes satisfiability.

A node can only be assigned into the new genome $\mathcal{G}^\prime$ when all its dependencies are satisfied.
At first, the \emph{addable} set $Q$ contains only nodes depending on input nodes---as these are the only nodes with satisfied dependencies.
However, the positional bias states that active nodes are likely to concentrate near input nodes.
Again, these active nodes have little or no computational nodes prior to them.
Hence, they have a higher probability of depending on input nodes compared to other nodes.
As a consequence, computational nodes near input nodes have a higher probability of being added into the \emph{addable} set early, as their dependencies are likely to satisfy early.
From this, it follows that nodes near input nodes are also more likely to be inserted into $\mathcal{G}^\prime$ early.
This leads to a lessoned but not completely mitigated positional bias.

\begin{figure}[t]%
    \centering
    \includegraphics[width=0.8\textwidth]{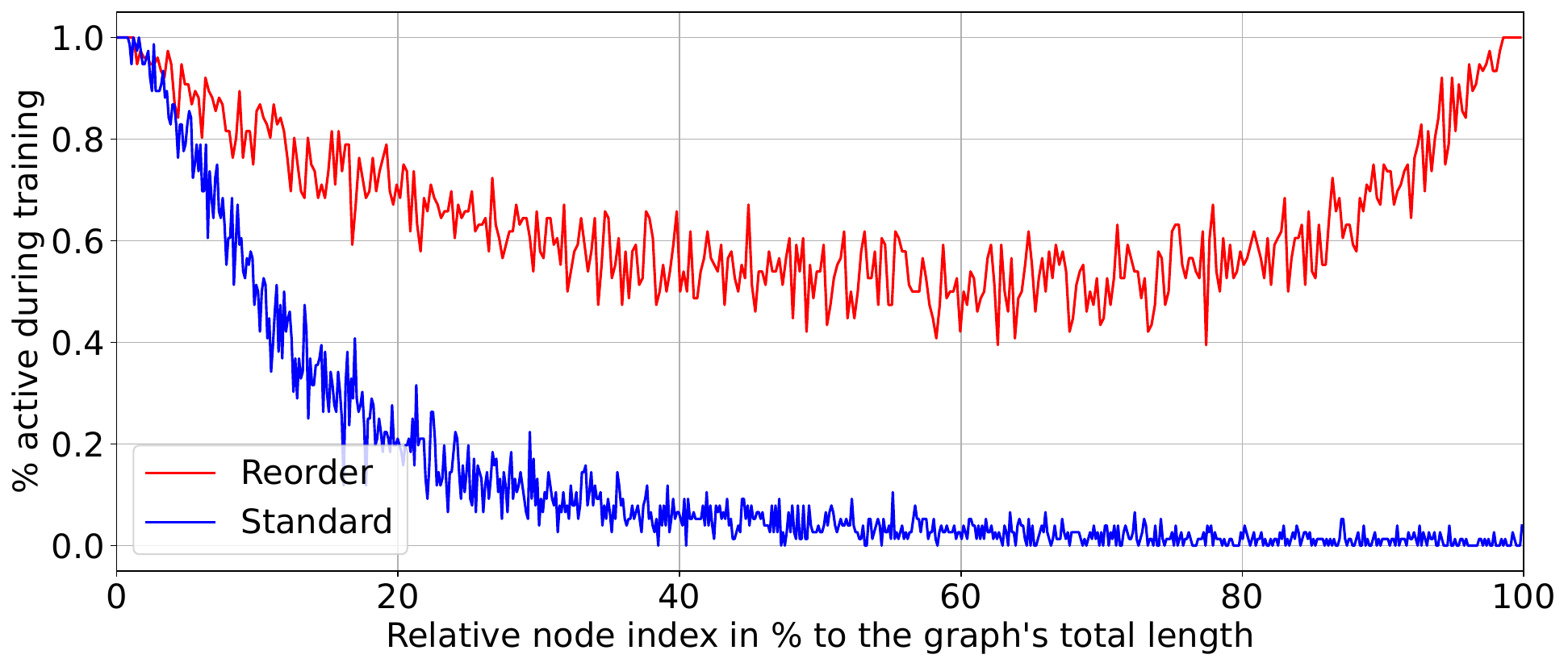}
    \caption{Distribution of active nodes for \Standard{} compared to \OGReorder{}.}
    \label{fig:reorder_bias}
\end{figure}

The lessened bias can be exemplarily visualized, as is seen in Figure~\ref{fig:reorder_bias}.
It shows the active node distribution of \Standard{} compared to \OGReorder{}.
Each distribution shows the averaged result of $75$ independent and aggregated solutions for the \emph{3-bit multiply} Boolean benchmark.
Again, the x-axis defines the position of a node in its graph, with the y-axis indicating its probability of being active.
The distribution for \OGReorder{} does not decrease as heavily as for \Standard{}.
Furthermore, the lowest probability for a node being active is around $40\%$ instead of almost zero.
However, there is still a visible decline of the nodes probability of becoming active.
While the positional bias is lessened for \OGReorder{}, it is not fully mitigated.

\subsection{Equidistant Reorder}
\label{subsec:reorder_e}
To fully avoid the positional bias, Cui et al.\cite{cui_ecta_reorder} introduced \emph{Equidistant-Reorder}.
Instead of depending on an addable set, active nodes are placed equidistantly apart in the grid first.
By enforcing an equidistant spacing, the positional bias is completely eliminated.
This tends to decrease the amount of computational nodes needed to train a CGP graph while also improving its fitness value.

To perform the Equidistant-Reorder operator on a genome, two different sets of nodes must be created.
Firstly, all active computational nodes are defined in the set $A$.
The second set $\tilde{A}$ contains all inactive computational nodes.
They are defined as:
\begin{align*}
    A &\coloneqq \{a_i \mid a_i \text{ is active computational node}, a_1 < a_2 < \cdots < a_n\}\\
    \tilde{A} &\coloneqq \{\tilde{a}_j \mid \tilde{a}_j \text{ is inactive computational node}, \tilde{a}_1 < \tilde{a}_2 < \cdots < \tilde{a}_{N-n}\}
\end{align*}
with $|A| \coloneqq n$ and $N =$ \# computational nodes.


Let $s \in \mathbb{N}_+$ be the starting index and $e \in \mathbb{N}_+$ be the last possible index of computational nodes in the genome.
Given $s$ and $e$, the new positions for active nodes can be given by set $L$, which is also described in Algorithm~\ref{algo:linspace}.
Here, $L$ is defined as equidistantly spaced numbers over the interval $[s, e]$, with $L \coloneqq \left\{\lfloor s + i \cdot \frac{e - s}{n}\rfloor \mid i = 1, \cdots, n \right\}$.
Given this formula, if $n=1$, then $\mathit{L} = \{e\}$.
If there exists only one active node, it will be placed at the end of the genome just before the output nodes.
As a result, the active node is able to mutate its connection to an arbitrary computational node without any restrictions.

Opposing to $L$, $\tilde{L}$ is the set of genome locations for inactive nodes, which is defined as $\tilde{L} \coloneqq \{s, s+1, \cdots, e\}~\setminus~\mathit{L} \coloneqq \{\tilde{l}_1, \cdots, \tilde{l}_{N-n}\}$ with $\tilde{l}_1 < \tilde{l}_2 < \cdots < \tilde{l}_{N-n}$.

Afterwards, a new genome $\mathcal{G}^\prime$ is initialized containing all input and output nodes from the original genome.
Again, input and output nodes will not be assigned new positions.
All active computational nodes are placed into their new positions in $\mathcal{G}^\prime$.
This is done by assigning $a_i \in \mathit{A}$ the position $l_i \in \mathit{L}$, for $i =1, \cdots, n$.
The same approach is used for inactive nodes $\tilde{a}_j \in \mathit{\tilde{A}}$, as they are placed in $\tilde{l}_j \in \mathit{\tilde{L}}$ for $j = 1, \cdots, N-n$.
As the order of computational nodes are not modified, the reordering does not change the semantic and no re-evaluation is required.

After inserting nodes, their connection genes must be corrected by changing their connection from the old genome's location to the new one.
However, some inactive nodes may now hold a connection to a node in front of them.
This can happen if they are connected to an active node, which gets a new position in front of it.
As a result, the graph is not feed-forward anymore, violating the representation of CGP.
In this case, a new connection to a previous node must be mutated to ensure the graph being feed-forward again.

The full pseudocode for a better understanding of this operator is depicted in Algorithm~\ref{algo:e_reorder}.
Furthermore, to improve readability, the Equidistant Reorder operator is called \EReorder{} in the following sections.

\begin{algorithm}[t]
    \caption{Lin-Space}\label{algo:linspace}
    \begin{algorithmic}[1]
        \Require start point $s$, end point $e$, number of evenly spaced values $n$
        \State $step \gets \frac{e - s}{n}$
        \State $S \gets \emptyset$
        \ForEach{$i = 1, \cdots, n$}
        \State $S \gets S \cup \lfloor s + i \cdot step\rfloor$\
        \EndFor
        \State \Return $S$
    \end{algorithmic}
\end{algorithm}

\begin{algorithm}[t]
    \caption{Equidistant Reorder}\label{algo:e_reorder}
    \begin{algorithmic}[1]
        \State $\mathit{A} \gets$ all active computational nodes
        \State $\mathit{\tilde{A}} \gets$ all inactive computational nodes
        \State Initialize new genome $\mathcal{G}^\prime$
        \State $s \gets |input\_nodes| + 1$
        \State $e \gets |input\_nodes| + |A| + |\tilde{A}|$
        \State $\mathit{L} \gets \text{Lin-Space}(s, e, |A|)$
        \State $\mathit{\tilde{L}} \gets \{s, s+1, \cdots, e\} \setminus \mathit{L}$
        
        \ForEach{$\left(a, l\right) \in (A, L)$}        
        \State     $G^\prime\left[l\right] \gets G\left[a\right]$
        \EndFor
        
        \ForEach{$\left(\tilde{a}, \tilde{l}\right) \in (\tilde{A}, \tilde{L})$}        
        \State    $G^\prime\left[\tilde{l}\right] \gets G\left[\tilde{a}\right]$
        \EndFor

        \ForEach{$node \in G^\prime$}        
        \ForEach{$connection$ belonging to $node$}        
        \State $update(connection)$
        \If{$connection$ points to the front}
        \State mutate$\left(connection\right)$\
        \EndIf
        \EndFor
        \EndFor
        
    \end{algorithmic}
\end{algorithm}

\subsection{Uniformly Distributed Reorder}
\label{subsec:reorder_uniform}
In the original work, \EReorder{} is able to outperform \OGReorder{} and \Standard{} in most cases~\cite{cui_ecta_reorder}.
However, by strictly enforcing an equidistant spacing, we believe that some useful structures might be destroyed.
As the order of operation \emph{for inactive nodes} are not conserved, they might have to mutate their connection to ensure a cycle-free graph.
This, in turn, might dissolve meaningful structures.

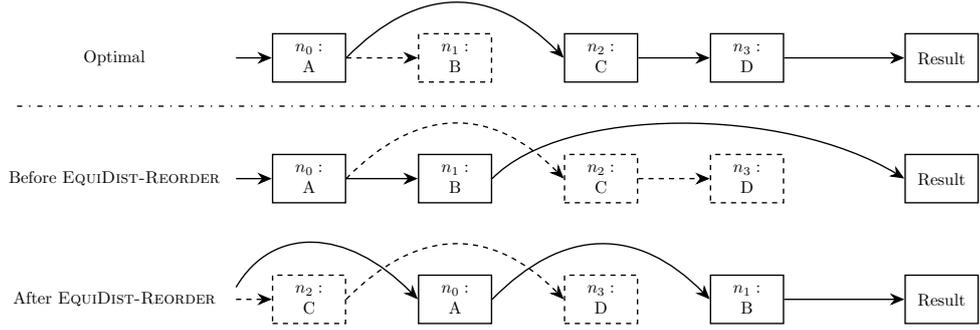
\begin{figure}[t]
    \centering
    \resizebox{\linewidth}{!}{%
        \begin{tikzpicture}
            [>={Stealth[width=2.5mm,length=3mm]},
            thick,
            cgpnode/.style={rectangle, draw, minimum width = 1.5cm, minimum height = 1cm, align=center}]
            \draw[-, loosely dashdotted] (-6, 1.5) to (14, 1.5 );
            
            \node () at (-4,2.5) [] {Optimal };
            \node (a0) at (0,2.5) [cgpnode] {$n_0:$ \\ A};
            \node (b0) at (3,2.5) [cgpnode, dashed] {$n_1:$ \\ B};
            \node (c0) at (6,2.5) [cgpnode] {$n_2:$ \\ C};
            \node (d0) at (9,2.5) [cgpnode] {$n_3:$ \\ D};
            \node (r0) at (13,2.5) [cgpnode] {Result};
            
            \draw[->] (-1.5, 2.5) to (a0.west);
            \draw[->, dashed] (a0.east) to (b0.west);
            \draw[->, ] (a0.east) .. controls ($(b0.north west) + (0, 1)$) and ($(b0.north east) + (0, 1)$) .. (c0.west);
            \draw[->, ] (c0.east) to (d0.west);
            \draw[->, ] (d0.east) to (r0.west);
            
            \node () at (-4,0) [] {Before \EReorder};
            \node (a) at (0,0) [cgpnode] {$n_0:$ \\ A};
            \node (b) at (3,0) [cgpnode] {$n_1:$ \\ B};
            \node (c) at (6,0) [cgpnode, dashed] {$n_2:$ \\ C};
            \node (d) at (9,0) [cgpnode, dashed] {$n_3:$ \\ D};
            \node (r) at (13,0) [cgpnode] {Result};
            
            \draw[->] (-1.5, 0) to (a.west);
            \draw[->] (a.east) to (b.west);
            \draw[->, dashed] (a.east) .. controls ($(b.north west) + (0, 1)$) and ($(b.north east) + (0, 1)$) .. (c.west);
            \draw[->, dashed] (c.east) to (d.west);
            \draw[->] (b.east) .. controls ($(c.north west) + (0, 1)$) and ($(d.north east) + (0, 1)$) .. (r.west);
            
            \node () at (-4, -2.5) [] {After \EReorder};
            \node (c2) at (0,-2.5) [cgpnode, dashed] {$n_2:$ \\ C};
            \node (a2) at (3,-2.5) [cgpnode] {$n_0:$ \\ A};
            \node (d2) at (6,-2.5) [cgpnode, dashed] {$n_3:$ \\ D};
            \node (b2) at (9,-2.5) [cgpnode] {$n_1:$ \\ B};
            \node (r) at (13,-2.5) [cgpnode] {Result};
            
            \draw[->, dashed] (-1.5, -2.5) to (c2.west);
            \draw[->] (b2.east) to (r.west);
            \draw[->] (a2.east) .. controls ($(d2.north west) + (0, 1)$) and ($(d2.north east) + (0, 1)$) .. (b2.west);
            \draw[->] (-1.5, -2.25) .. controls ($(c2.north west) + (0, 1)$) and ($(c2.north east) + (0, 1)$) .. (a2.west);
            \draw[->, dashed] (c2.east) .. controls ($(a2.north west) + (0, 1)$) and ($(a2.north east) + (0, 1)$) .. (d2.west);
        \end{tikzpicture}
    }
    \caption{
        Thought experiment illustrating limitations in \EReorder{}. By reordering the nodes, the optimal solution is harder to obtain. Before reordering nodes, the node $Result$ must mutate its connection from node $B$ to $D$. After reordering, more changes must be made to obtain the optimal solution.
    }
    \label{fig:ereorder_problem}
\end{figure}

Such effects of \EReorder{} can be examined using a thought experiment, shown in Figure~\ref{fig:ereorder_problem}.
The optimal ordering for this program would be obtained by mutating one connection, from $B \rightarrow Result$ to $D \rightarrow Result$.
However, by reordering nodes, such structures get destroyed. 
With node $A$ in front of node $C$, it is not possible to obtain the optimal solution by changing only one connection.
While the optimal solution is still possible to achieve, it is much harder after the reorder operator.
The reason is that structures must be reinvented, which involves more changes in the genotype. 

Motivated by this thought experiment, we introduce an operator similar to \EReorder{} but without enforcing an equidistant spacing: \emph{Uniformly Distributed Reorder} (\UDReorder{}).
Here, same algorithm described in Section~\ref{subsec:reorder_e} applies to \UDReorder{}.
The only difference lies in obtaining the new positions for active nodes, given by the set $L$.
Instead of using Algorithm~\ref{algo:linspace}, we now sample from a continuous uniform distribution over the range $[s, e]$, with $s$ being the starting index and $e$ being the last index of computational nodes in the grid.
This means, $L\sim \mathcal{U}_{\left[s, e\right]}$.

\subsection{Negative Positional Bias based Reorder}
\label{subsec:reorder_negative_bias}
Another novel operator, which reorders and shuffles the genotype, will be called \emph{Negative Positional Bias based Reorder} (\NPBReorder{}).
The idea of \NPBReorder{} is to move \emph{all} active nodes to the end of the genotype just before output nodes. 
Let $n$ be the number of active nodes and $e$ be the last possible index for computational nodes in a genome with $n, e \in \mathbb{N}_+$. Then, all active nodes are moved in their respective ordering to the positions $\left\{e-n+1, e-n + 2, \cdots, e\right\}$.
Concerning the inactive nodes, they are moved before all active nodes.
Again, as is also the case for \EReorder{}, all illegal connections must be mutated anew. 
For \NPBReorder{}, Algorithm~\ref{algo:e_reorder} can be used to describe its approach.
The algorithm itself does not change with the sole exception of obtaining $L$, which is the set of new positions for active nodes.
Instead of using Algorithm~\ref{algo:linspace}, we define $L\coloneqq\left\{e-n+1, e-n + 2, \cdots, e\right\}$.

By using this operator, we believe that it will have two different effects on the genotype.
At first, active nodes now have a very high chance of mutating a connection to inactive nodes.
Positive effects can be argued with CGP's \emph{neutral genetic drift}, which describes the positive influence of \emph{inactive nodes}.
This should allow drastic changes in the phenotype during the following training iterations, which can play an important role in escaping local optima~\cite{neutral_drift_3,neutral_drift2,neutral_drift}.
The other effect on the genotype is the mutation of new connections, which some inactive nodes must perform.
For example, given the optimally found hyperparameters for \Standard{} on the \emph{3-bit multiply} Boolean benchmark averaged over 20 independent and randomly seeded runs, a mean of 863 out of 1,000 nodes were \emph{inactive}.
For \emph{these inactive nodes}, a mean of 409 nodes had at least one connection to an active node.
Because these inactive nodes are connected to an active node, this indicates that about half of all inactive nodes mutate due to \NPBReorder{}, which introduces further changes to the genotype.
Nonetheless, deliberately mutating inactive nodes may improve CGP's performance for \Standard{}, as was found by Turner and Miller~\cite{neutral_drift} or Cui et al.~\cite{Cui2022}.
Furthermore, according to Kaufmann and Kalkreuth~\cite{on_the_param}, mutating a connection gene is the most meaningful change in the genotype of a CGP graph without extensions---as is the case for \Standard{}.
They show that changing connection genes can improve the fitness at most.
To summarize, by introducing drastic changes in the genotype, it might help to efficiently explore the search space, and the escape of local optima.

We believe that Boolean benchmarks might benefit the most from \NPBReorder{}.
According to Vasicek~\cite{Vasicek2018}, the fitness landscape of such benchmarks are \emph{deceptive}.
In this context, this means that a multitude of different solutions lead to the same fitness value.
Hence, by enforcing a drastic change in the phenotype due to \NPBReorder{}, more regions in the fitness landscape might be explored compared to \Standard{}.

However, reordering active nodes toward output nodes after each iteration might hinder the evolutionary search process---as the genotype is drastically changing.
This is why we also introduce a hyperparameter $p_{reorder} \in [0, 1]$, which describes the probability of \NPBReorder{} being executed.
By varying the frequency of reorder operations, an equilibrium might be found between exploiting the current search space area and exploring new regions.

\subsection{Left-Skewed Distribution Reorder}
\label{subsec:reorder_left_skewed}
Similarly to \EReorder{}, enforcing a strict placement of nodes might greatly increase the difficulty to use existing genotypical structures.
This is why we also include a probability based approach to \NPBReorder{} called \emph{Left-Skewed Distribution Reorder} (\LSDReorder{}).

\begin{figure}[t]%
    \centering
    \includegraphics[width=0.7\textwidth]{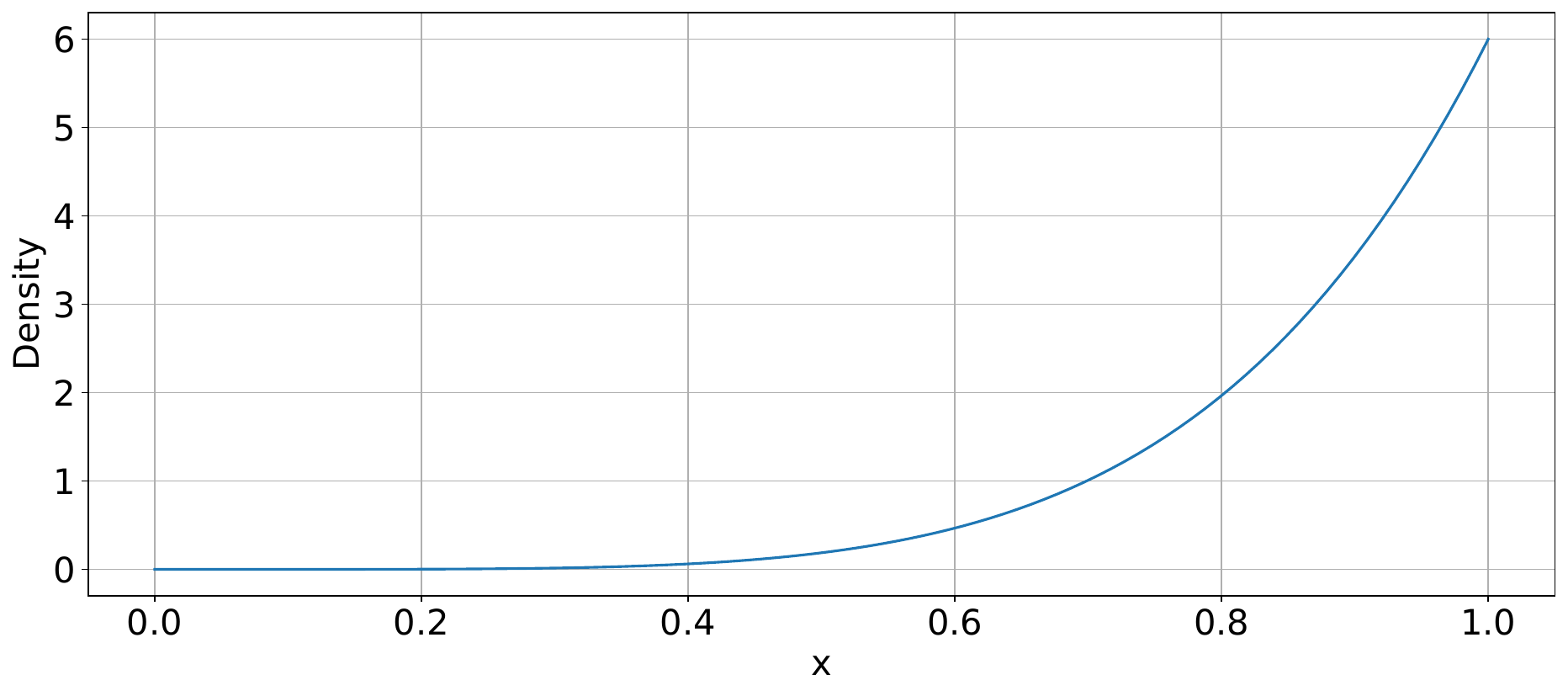}
    \caption{Probability density function of the beta distribution with $\alpha = 6$ and $\beta=1$.}
    \label{fig:pdf_beta}
\end{figure}

For this operator, active nodes are reordered according to a left-skewed beta distribution with $\alpha = 6$ and $\beta=1$. 
Again, Algorithm~\ref{algo:e_reorder} can be reused to better visualize this operator.
The majority of the algorithm does not change with the exception of the set of new active node positions $L$.
It is sampled by the aforementioned beta distribution $L \sim Beta(6, 1)$.
To better visualize this specific distribution, its probability density function is also shown in Figure~\ref{fig:pdf_beta}.
Again, similarly to \NPBReorder{}, the \LSDReorder{} operator also includes a hyperparameter $p_{reorder} \in [0, 1]$ to describe the probability of a reorder occurring.

\subsection{Time Complexity}
The original authors of \OGReorder{} show a time complexity of $\mathcal{O}(M \cdot N)$, where $a \in \mathbb{N}_+$ is the arity of the nodes and $N \in \mathbb{N}_+$ is the number of computational nodes of the considered CGP graph~\cite{dag_extension}.
In their previous paper, Cui et al.~\cite{cui_ecta_reorder} showed that \EReorder{} also needs $\mathcal{O}(a \cdot N)$ time, which is the same as the original counterpart.

To obtain the time complexity for \UDReorder{}, \NPBReorder{} and \LSDReorder{}, their similarities and differences must be compared first.
The reason is that they only differ in obtaining the set of new positions for active nodes.

Getting the set of all active and inactive computational nodes takes $\mathcal{O}(a \cdot N)$ time by following the connections from output nodes first backwards to input nodes.
For $\tilde{L}$, a time of $\mathcal{O}(N)$ is needed to calculate the difference of two sets.
The placing of active and inactive nodes into a new genome takes only $\mathcal{O}(N)$ time, as this only requires iterating through the sets of active and inactive nodes and their respective new positions once.
For the procedure of updating the connections of every inactive computational node, each one must be visited to potentially update its $c$ connection genes.
This takes $\mathcal{O}(c \cdot N)$ time.
Please note, $a \equiv c$, as the highest arity $a$ is also the number of connection genes $c$.

To obtain the time complexity for $L$, two cases must be distinguished.
For \NPBReorder{}, a set of numbers in a given range is generated. 
This can be done in $\mathcal{O}(N)$.
Considering \UDReorder{} and \LSDReorder{}, pseudo-random numbers must be generated from a predefined distribution.
Here, initializing the distribution might be complex but is only done once before the start of the training process.
Simply sampling $N$ values from such distributions, however, can thus be achieved in amortized $\mathcal{O}(N)$.

Hence, \UDReorder{}, \NPBReorder{} and \LSDReorder{} need $\mathcal{O}(a \cdot N)$ time, which is the same as \OGReorder{} and \EReorder{}.

\section{Evaluation of Different Reorder Strategies}
\label{sec:evaluation}

To gauge the effectiveness of our proposed methods, we conducted an empirical study. 
We describe our experimental design and attempt to answer the following questions:

\begin{description}
    \item[Q1:] Which CGP variant needs, considering the learning task, less training iterations and/or has the better fitness value:
    \begin{enumerate}
        \item \Standard{}
        \item \OGReorder{}
        \item \EReorder{}
        \item \UDReorder{}
        \item \NPBReorder{}
        \item \LSDReorder{}
    \end{enumerate}
    \item[Q2:] Do these variants behave differently regarding their convergence behaviour?
\end{description}

\subsection{Experimental Design}
In this section, we describe CGP's configuration and the used benchmarks.
Afterwards, a brief introduction into Bayesian comparison of models to evaluate and rank different CGP configurations is given.
These models are used to compare multiple CGP configurations, which we use for our statistical analysis. 

\subsubsection{Bayesian Data Analysis}
\label{subsubsec:bayesian}

To ensure a fair comparison of extensions and a qualitatively sound evaluation, multiple configurations were investigated for each CGP version.
Thus, each hyperparameter configuration has to be ranked to find the best solution.
In our experiments, for all Boolean benchmarks we only examine the number of training iterations until a solution is found.
The fitness value is considered regarding symbolic regression benchmarks.
These numbers cannot be negative.
Hence, other common distributions such as Student's $t$ distributions can not be expected to model the data well~\cite{kruschke}.
This is why we performed a \emph{Bayesian data analysis}---we utilized the Python library \emph{cmpbayes}~\cite{cmpbayes} for all statistical models---for the posterior distributions of our results.
The model to compare the algorithms is based on the Plackett-Luce model described by Calvo et al.~\cite{calvo}.
It allows the computation of a set of ranked options by estimating the probabilities of each of the options to be the one with the highest rank.

Additionally, for each CGP version and their respective best hyperparameter set found, we report the 95\,\% \emph{highest posterior density intervals} (HPDI) of the distribution of $\mu_{config}$, where $\mu_{config}$ is a random variable corresponding to the respective \emph{performance measurement}.
At that, the distribution of $\mu_{config}$ is estimated by the gamma distribution--based model for comparing non-negative data from \emph{cmpbayes}~\cite{cmpbayes}.
Please note, a 95\,\% HPDI interval $[l, u]$ can be read as $p(l~\leq~\mu_{config}~\leq u)~=~95\,\%$.
This means, the probability of the algorithms results lying between the bounds $l$ and $u$ has a probability of 95\,\%.

Furthermore, prior sensitivity analyses were conducted prior to ensure the robustness of all models.
As they always display similar results, robust and meaningful models are implicated.
Finally, please note that \emph{cmpbayes} uses Markov Chain Monte-Carlo sampling to obtain its distributions.
Therefore, the usual checks to ensure convergence and well-behavedness (trace plots, posterior predictive checks, $\hat{R}$ values, effective sample size) were performed.
For more information regarding the models, we refer to Kruschke and P\"atzel~\cite{kruschke,cmpbayes}.

\subsubsection{Evolutionary Algorithms and Configuration}
As is standard for most CGP configurations, an elitist $\left(\mu + \lambda \right)$ evolutionary strategy (ES) with $\mu = 1$ and $\lambda = 4$ is used~\cite{neutral_drift,cgp_book}.
Furthermore, CGP usually does not rely on crossover, as it generally does not improve CGP's performance~\cite{Husa2018}.
This is why we also do not use one.
Considering our mutation operator, we use \emph{Single}~\cite{single_mutation}.
It has the benefit that it does not need a mutation probability and achieves similar results compared to a standard point mutation.
This operator works by mutating random nodes until an active node was mutated, which enforces a measurable change in the phenotype.

One major hyperparameter which must be optimized for all CGP versions and benchmarks is its number of computational nodes $N$.
Therefore, we investigated $N~\in~\left\{50, 100, 150, \cdots, 2000\right\}$ and each $N$ was tested $20$ times with independent repetitions and different random seeds.
If a probability for reorder $p_{reorder}$ is needed, grid search was used.
In addition to $N$, a $p_{reorder}~\in~\left\{0.0, 0.1, 0.2, \cdots, 1.0\right\}$ was examined.
To find the best hyperparameter setting, we used the Plackett-Luce model described by Calvo et al.~\cite{calvo}.

\subsubsection{Benchmarks}
To evaluate the methods, benchmarks from two different problem domains are used: Boolean and symbolic regression benchmarks.

\paragraph{Boolean Benchmarks}
We used four Boolean benchmarks problems: 3\emph{-bit Parity}, 16-4\emph{-bit Encode}, 4-16\emph{-bit Decode} and 3\emph{-bit Multiply}.
In the following, we will call these \emph{Parity}, \emph{Encode}, \emph{Decode} and \emph{Multiply}, respectively.
Parity is regarded as too easy by the Genetic Programming community~\cite{datasets_which}.
Nevertheless, it is commonly used as a benchmark in the literature~~\cite{neutral_drift2,on_the_param,ex_parity1}.
Hence, we also included it in our evaluations for ease of comparison.
Encode and Decode are problems with different input and output sizes (16 inputs and 4 outputs, and vice versa).
At last, Multiply is a comparatively hard problem~\cite{multiply_benchmark}, and recommended by White et al.~\cite{datasets_which}.
In addition, all four benchmarks were used to evaluate \OGReorder{}~\cite{dag_origin,dag_extension} and \EReorder{}~\cite{cui_ecta_reorder}, which is also why we utilized them for better comparison.

As we employed four Boolean benchmark problems, we also trained them with the standard function set for these problems.
They contain the Boolean operators \emph{AND}, \emph{OR}, \emph{NAND} and \emph{NOR}.
These benchmarks also lead to a standard fitness function, which is defined by the ratio of correctly mapped inputs.
Let $f:\{0, 1\}^m \rightarrow \{0, 1\}^n$ be a correct Boolean mapping for $m \in \mathbb{N}_+$ inputs and $n \in \mathbb{N}_+$ outputs.
Then, the fitness of an individual $g: \{0, 1\}^m \rightarrow \{0, 1\}^n$, which relates to the learning task $f$, is defined as follows:
\begin{equation}
    \frac{\lvert\{x \in \{0, 1\}^m \mid f(x)=g(x)\}\rvert}{\lvert\{0, 1\}^m\rvert}
\end{equation}

The goal for Boolean benchmarks is to achieve a solution which is able to correctly map all inputs. 
Thus, each benchmark runs on an unlimited budget and we report the \emph{number of training iterations until a solution is found (I2S)}.
As four children are generated and evaluated in each iteration, the spent budget would be four times the I2S.

\paragraph{Symbolic Regression Benchmarks}
In terms of symbolic regression benchmarks, we adhered to the recommendations from the GP community~\cite{datasets_which} and previous works~\cite{Kalkreuth2020}.
Again, four different benchmarks were used: \emph{Nguyen-7}, \emph{Koza-3}, \emph{Pagie-1} and \emph{Keijzer-6}.
Their functions are shown in Table~\ref{tab:regression_benchmarks}.

\begin{table}[t]
    \caption{Symbolic regression benchmarks used. $U[a, b, c]$ means that $c$ uniform random samples are drawn from $a$ to $b$, inclusive. $E[a, b, c]$ defines a grid of points from $a$ to $b$, with $c$ being the spacing.}
    \label{tab:regression_benchmarks}
    {\renewcommand{\arraystretch}{1.2}
        \begin{tabular}{@{}lllll@{}}
            \toprule
            Name & Variables & Equation & Training Set & Test Set \\
            \midrule
            Nguyen-7 & 1 & $\ln\left(x + 1\right) + \ln\left(x^2 + 1\right)$ & $U[0, 2, 20]$ & None \\
            Koza-3 & 1 & $x^6 - 2 \cdot x^4 + x^2$ & $U[-1, 1, 20]$ & None \\
            Pagie-1 & 2 & $\frac{1}{1 - x^{-4}} + \frac{1}{1 - y^{-4}}$ & $E[-5, 5, 0.4]$ & None \\
            Keijzer-6 & 1 & $\sum_{i}^{x}\frac{1}{i}$ & $E[1, 50, 1]$ & $E[1, 120, 1]$ \\
            \botrule
        \end{tabular}
    }
\end{table}

The function set consists of eight mathematical functions: addition, subtraction, multiplication, protected division, sine, cosine, natural logarithm and the exponential function.
As for the fitness function, the \emph{mean absolute error} over the whole benchmark with $n$ entries was used:
\begin{equation}
    \frac{1}{n} \sum_{i = 1}^{n}|y_i - x_i|
\end{equation}
with $y_i \in \mathbb{R}$ being the prediction of a model and $x_i \in \mathbb{R}$ being the corresponding true value, for $i = 1, \cdots, n$.

In this setting, an algorithm is classified as converged when the fitness value becomes less than $0.01$.
Furthermore, each CGP variant is given $5 \cdot 10^5$ training iterations per run.
As we employ a $(1+4)$-ES, each of the four children have to be evaluated during a single training iteration.
This results in an upper limit for the budget of $2 \cdot 10^6$ fitness evaluations for symbolic regression benchmarks.

\subsection{Performance on Boolean Benchmarks}
\label{subsec:performance_boolean}
\begin{table}[t]
    \caption{Results on Boolean benchmarks for each CGP variant and its corresponding best configuration found. Here, \emph{Active} is the mean number of active nodes; \emph{Nodes} is the number of computational nodes used; and $p(best)$ is the probability of the solution having the best test fitness value. Algorithms are sorted from best to worst in descending order according to $p\left(best\right)$. }
    \label{tab:results_boolean}
    \begin{tabular}{@{}lcccccc@{}}
        \toprule
        Variant & $Mean(I2S)$ & HPDI  & Active & Nodes  & $p_{reorder}$ & $p\left(best\right)$\\
        \midrule
        \multicolumn{7}{c}{\textbf{Parity}}  \\
        \OGReorder{} & $343$ & $[281, 419]$ & 46.5 & 600 & --- & $0.22$ \\
        \Standard{} & $406$ & $[329, 503]$ & 30.5 & 200 & --- & $0.18$ \\
        \UDReorder{} & $423$ & $[349, 515]$ & 44.9 & 550 & --- & $0.18$ \\
        \LSDReorder{} & $402$ & $[324, 498]$ & 42.3 & 500 & $1.0$ & $0.15$ \\
        \EReorder{} & $464$ & $[377, 566]$ & 37.2 & 400 & --- & $0.14$ \\
        \NPBReorder{} & $545$ & $[427, 693]$ & 45.9 & 700 & $0.5$ & $0.13$ \\
        \midrule
        \multicolumn{7}{c}{\textbf{Encode}}  \\
        \UDReorder{} & $6,178$ & $[5,380, 7,086]$ & 78.8 & 350 & --- & $0.20$ \\
        \LSDReorder{} & $6,466$ & $[5,646, 7,385]$ & 69.3 & 250 & $0.4$ & $0.18$ \\
        \NPBReorder{} & $6,413$ & $[5,696, 7,198]$ & 66.0 & 200 & $0.8$ & $0.17$ \\
        \Standard{} & $6,544$ & $[5,627, 7,569]$ & 82.4 & 400 & --- & $0.16$ \\
        \OGReorder{} & $6,695$ & $[5,722, 7,827]$ & 94.9 & 500 & --- & $0.16$ \\
        \EReorder{} & $7,699$ & $[6,591, 9,005]$ & 71.3 & 250 & --- & $0.13$ \\
        \midrule
        \multicolumn{7}{c}{\textbf{Decode}}  \\
        \LSDReorder{} & $15,275$ & $[13,859, 16,905]$ & 166.8 & 550 & $0.8$ & $0.21$ \\
        \Standard{} & $15,432$ & $[13,913, 17,100]$ & 161.3 & 500 & --- & $0.20$ \\
        \EReorder{} & $16,177$ & $[14,274, 18,319]$ & 162.9 & 500 & --- & $0.19$ \\
        \UDReorder{} & $16,878$ & $[15,221, 18,711]$ & 173.8 & 600 & --- & $0.16$ \\
        \OGReorder{} & $17,069$ & $[15,139, 19,210]$ & 152.4 & 450 & --- & $0.14$ \\
        \NPBReorder{} & $18,470$ & $[16,844, 20,229]$ & 129.8 & 300 & $0.6$ & $0.11$ \\
        \midrule
        \multicolumn{7}{c}{\textbf{Multiply}}  \\
        \NPBReorder{} & $87,246$ & $[72,937, 103,316]$ & 125.2 & 750 & $0.9$ & $0.21$ \\
        \EReorder{} & $88,291$ & $[77,922, 99,814]$ & 91.1 & 350 & --- & $0.19$ \\
        \LSDReorder{} & $94,940$ & $[82,322, 109,039]$ & 105.2 & 550 & $0.5$ & $0.18$ \\
        \UDReorder{} & $100,378$ & $[84,129, 119,600]$ & 90.4 & 350 & --- & $0.18$ \\
        \OGReorder{} & $103,685$ & $[89,632, 119,653]$ & 101.2 & 500 & --- & $0.16$ \\
        \Standard{} & $145,299$ & $[121,919, 172,348]$ & 117.2 & 750 & --- & $0.08$ \\
        \botrule
    \end{tabular}
\end{table}

We report for each CGP variant its best configuration with respect to our hyperparameter search.
To generate our results, each configuration was run $75$ times with independent repetitions and different random seeds.
Furthermore, we state the mean number of I2S, its HPDI, number of (active) nodes, and $p_{reorder}$ if applicable.
Additionally, we report the probability of one solution being the best out of the six, in accordance to the Bayesian model defined in Section~\ref{subsubsec:bayesian}.
The summaries of our results on Boolean benchmarks can be seen in Table~\ref{tab:results_boolean}.

For all benchmarks, at least one reorder variant is always better than a standard CGP version.
However, there is no reorder operator which behaves best for all benchmarks.
Interestingly, for Parity, \OGReorder{} is preferred and the other shuffling methods do not lead to better results.
However, these results should be taken with reservations as the different I2S are close and it is deemed as too easy.

Considering the other benchmarks, \NPBReorder{} or \LSDReorder{} might be preferable in this setting.
\LSDReorder{} is at the second place for Encode and the best option for Decode, while \NPBReorder{} seems to be the best variant for Multiply.
Their effectiveness might be explained due to their high amount of genetic drift and phenotypical changes, which they introduce each time.
Combined with the deceptive fitness landscape mentioned in Section~\ref{subsec:reorder_negative_bias}, this might be a reason why such reorder operators might be favourable.
In addition, a high $p_{reorder}$ value is preferred for almost all instances, which supports the claim of high exploration being favourable.

Another interesting insight is that \EReorder{} and \UDReorder{} do not perform similarly, and the same observation can be seen for \NPBReorder{} and \LSDReorder{}.
As their shuffling methods are very similar, we would expect a similar fitness as well.
However, this is not the case for these four benchmarks.
Nevertheless, some trends can be seen.
For both harder benchmarks---Decode and Multiply---\EReorder{} seems to be preferable over \UDReorder{}, while the opposite can be said about Parity and Encode: Here, \UDReorder{} leads to better results than \EReorder{}.

One major caveat of \NPBReorder{} and \LSDReorder{}, however, is their need for an additional hyperparameter $p_{reorder}$.
This value differs for each CGP variant and benchmark, which is why no universal recommendation for $p_{reorder}$ can be given.
As a result, this hyperparameter should be optimized for each problem statement, which leads to additional complexity and resources needed.
This leads to a trade-off, as CGP's performance increases slightly but with a higher complexity for its hyperparameter search.


\subsection{Performance on Symbolic Regression Benchmarks}
\label{subsec:performance_regression}

\begin{sidewaystable}
    \caption{Results on symbolic regression benchmarks for each CGP variant and its corresponding best configuration found. Here, \emph{Active} is the mean number of active nodes; \emph{Nodes} is the number of computational nodes used; \emph{SR} is the success rate; and $p(best)$ is the probability of the solution having the best test fitness value. Both $f(train)$ and $f(test)$ depict the respective mean fitness values. Algorithms are sorted from best to worst in descending order according to $p\left(best\right)$}\label{tab:results_regression}
    \begin{tabular*}{\textheight}{@{\extracolsep\fill}llccccccccc@{}}
        \toprule
        Benchmark & Variant & $Mean(I2S)$ & HPDI  & Active & Nodes  & $p_{reorder}$ & SR & $f(train)$ & $f(test)$ & $p\left(best\right)$\\ 
        \midrule
        \multirow{6}{*}{\textbf{Keijzer-6}} &
        \EReorder{} & $23$ & $[16, 32]$ & 11.6 & 50 & --- & $1.0$ & $0.001$ & $0.001$ & $0.28$ \\
        & \OGReorder{} & $25$ & $[18, 35]$ & 12.2 & 50 & --- & $0.96$ & $0.002$ & $0.028$ & $0.24$ \\
        & \LSDReorder{} & $28$ & $[20, 38]$ & 11.9 & 50 & $1.0$ & $1.0$ & $0.001$ & $0.001$ & $0.17$ \\
        & \NPBReorder{} & $37$ & $[26, 54]$ & 14.8 & 100 & $0.4$ & $1.0$ & $0.002$ & $0.001$ & $0.15$ \\
        & \UDReorder{} & $41$ & $[32, 53]$ & 24.4 & 300 & --- & $0.96$ & $0.004$ & $0.002$ & $0.10$ \\
        & \Standard{} & $480$ & $[312, 733]$ & 13.3 & 150 & --- & $0.92$ & $0.005$ & $0.025$ & $0.06$ \\
        \midrule
        \multirow{6}{*}{\textbf{Koza-3}} &
        \LSDReorder{} & $20,339$ & $[11,916, 34,454]$ & 24.8 & 150 & $0.7$ & $1.0$ & $0.006$ & $0.031$ & $0.22$ \\
        & \EReorder{} & $11,243$ & $[6,888, 18,108]$ & 20.8 & 100 & --- & $1.0$ & $0.006$ & $0.045$ & $0.19$ \\
        & \NPBReorder{} & $9,370$ & $[5,813, 15,051]$ & 49.0 & 700 & $0.2$ & $1.0$ & $0.007$ & $0.037$ & $0.17$ \\
        & \OGReorder{} & $14,302$ & $[8,676, 23,174]$ & 30.1 & 250 & --- & $1.0$ & $0.007$ & $0.023$ & $0.16$ \\
        & \Standard{} & $18,430$ & $[11,651, 28,939]$ & 28.1 & 650 & --- & $0.98$ & $0.009$ & $0.018$ & $0.14$ \\
        & \UDReorder{} & $18,544$ & $[11,183, 30,263]$ & 41.6 & 600 & --- & $0.98$ & $0.008$ & $0.028$ & $0.12$ \\
        \midrule
        \multirow{6}{*}{\textbf{Nguyen-7}} &
        \NPBReorder{} & $49,447$ & $[32,829, 73,519]$ & 33.4 & 250 & $0.4$ & $0.98$ & $0.009$ & $0.023$ & $0.19$ \\
        & \UDReorder{} & $55,563$ & $[36,527, 83,843]$ & 40.6 & 550 & --- & $0.94$ & $0.009$ & $0.829$ & $0.18$ \\
        & \EReorder{} & $54,855$ & $[35,814, 82,851]$ & 48.4 & 650 & --- & $0.96$ & $0.009$ & $0.906$ & $0.17$ \\
        & \LSDReorder{} & $59,022$ & $[39,292, 87,894]$ & 47.6 & 750 & $0.6$ & $0.96$ & $0.009$ & $0.033$ & $0.16$ \\
        & \Standard{} & $126,498$ & $[85,566, 182,333]$ & 33.5 & 700 & --- & $0.88$ & $0.011$ & $0.078$ & $0.10$ \\
        & \OGReorder{} & $73,330$ & $[47,996, 109,793]$ & 34.3 & 300 & --- & $0.92$ & $0.01$ & $0.034$ & $0.02$ \\
        \midrule
        \multirow{6}{*}{\textbf{Pagie-1}} &
        \EReorder{} & $352,345$ & $[271,198, 442,921]$ & 63.1 & 350 & --- & $0.43$ & $0.034$ & $0.034$ & $0.22$ \\
        & \OGReorder{} & $380,985$ & $[299,196, 471,922]$ & 63.4 & 350 & --- & $0.29$ & $0.036$ & $0.036$ & $0.19$ \\
        & \NPBReorder{} & $385,752$ & $[297,161, 485,266]$ & 63.9 & 450 & $1.0$ & $0.27$ & $0.047$ & $0.047$ & $0.19$ \\
        & \LSDReorder{} & $405,887$ & $[329,290, 493,721]$ & 56.3 & 300 & $0.8$ & $0.24$ & $0.049$ & $0.049$ & $0.15$ \\
        & \Standard{} & $491,071$ & $[459,092, 500,000]$ & 52.7 & 700 & --- & $0.02$ & $0.046$ & $0.046$ & $0.15$ \\
        & \UDReorder{} & $430,564$ & $[357,388, 500,000]$ & 47.1 & 150 & --- & $0.20$ & $0.065$ & $0.065$ & $0.10$ \\
        \botrule
    \end{tabular*}
\end{sidewaystable}

Considering the results on the symbolic regression benchmarks, they are shown in Table~\ref{tab:results_regression}.
Similar to Boolean benchmarks, each configuration was run $75$ times with independent repetitions and different random seeds.
Again, we list the CGP variant and their corresponding mean I2S, mean number of active and computational nodes, the probability of reorder (if applicable), success rate, train- and test fitness values, and the probability of the configuration being the best.
For the success rate, we report the fraction of solutions that achieved a training fitness value of less than $0.01$.
The probability of one solution being the best is calculated via the Bayesian model defined in Section~\ref{subsubsec:bayesian}.
The models and their corresponding probabilities were calculated on each configuration's test fitness values.
In addition, the Keijzer-6 benchmark has very low fitness and HPDI values, indicating that it is very easy for CGP to solve it quickly.
Thus, the results obtained from this benchmark should be met with scepticism.

Similarly to the Boolean benchmark results, there is no CGP variant that is always the best one.
However, applying a reorder method is always preferable over \Standard{}, as almost all configurations need less I2S while having a higher success rate.

Considering the different shuffling methods, \EReorder{} is the only one always under the top three CGP variants.
In contrast to that, \UDReorder{}---which has a similar reorder methods compared to \EReorder{}---is performing worse in three cases.
Similarly, \NPBReorder{} and \LSDReorder{} show different fitness and HPDI values, a behaviour that can also be observed for Boolean benchmarks.
Another similarity to Boolean benchmarks is the choice of $p_{reorder}$, as the optimally found values differ again for each benchmark and configuration.
As a result, no general value for $p_{reorder}$ can be proposed.
This, again, implies a trade-off between better performance but an additional hyperparameter to optimize.

What is also interesting to see for the Koza-3 benchmark is the ranking of algorithms and their I2S.
All top three variants have a success rate of one, which means that a solution was always found.
However, the best solution needs, on average, about double the training iterations than \EReorder{} and \NPBReorder{}.

\subsection{Convergence Behaviour}
\label{subsec:conv_behaviour}

\begin{figure}[t]%
    \centering
    \includegraphics[width=0.49\textwidth]{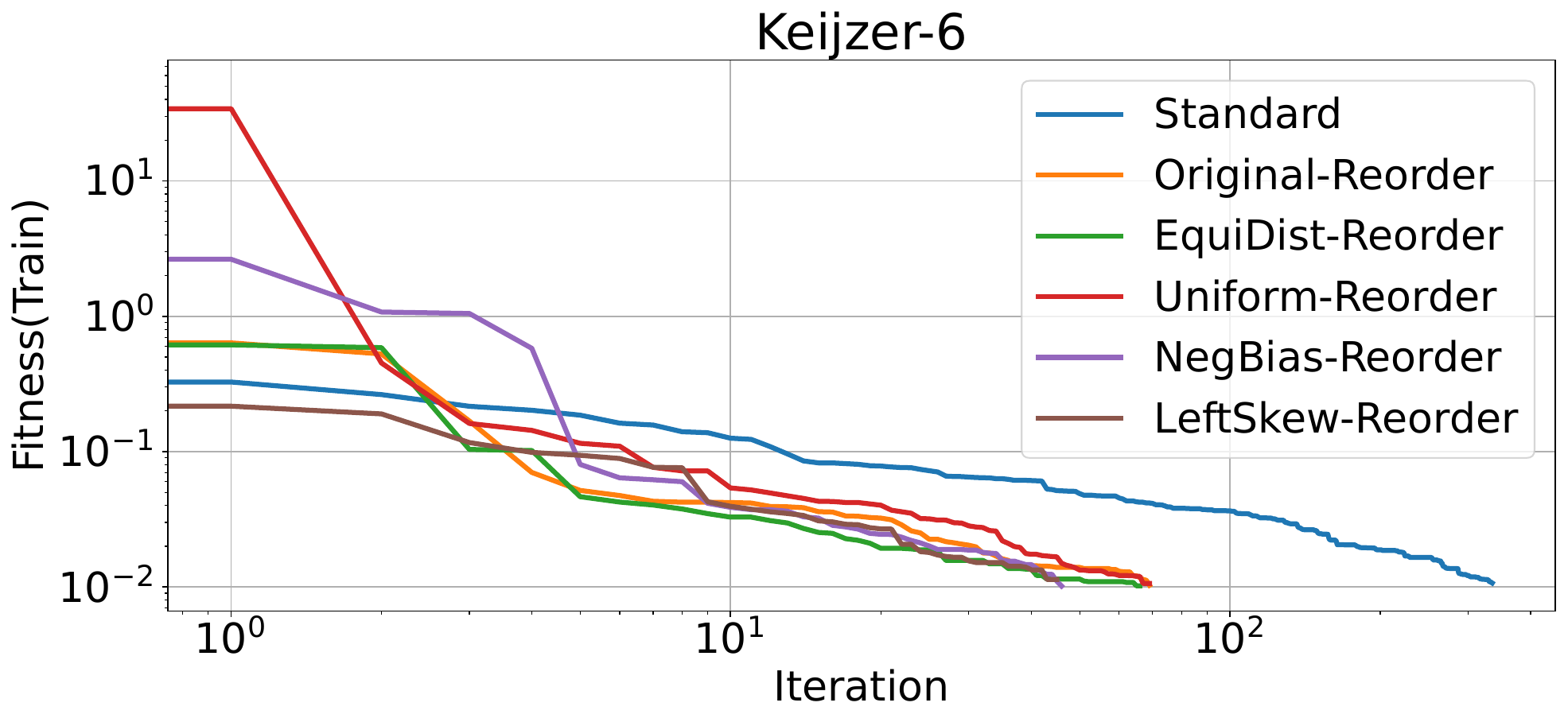}
    \includegraphics[width=0.49\textwidth]{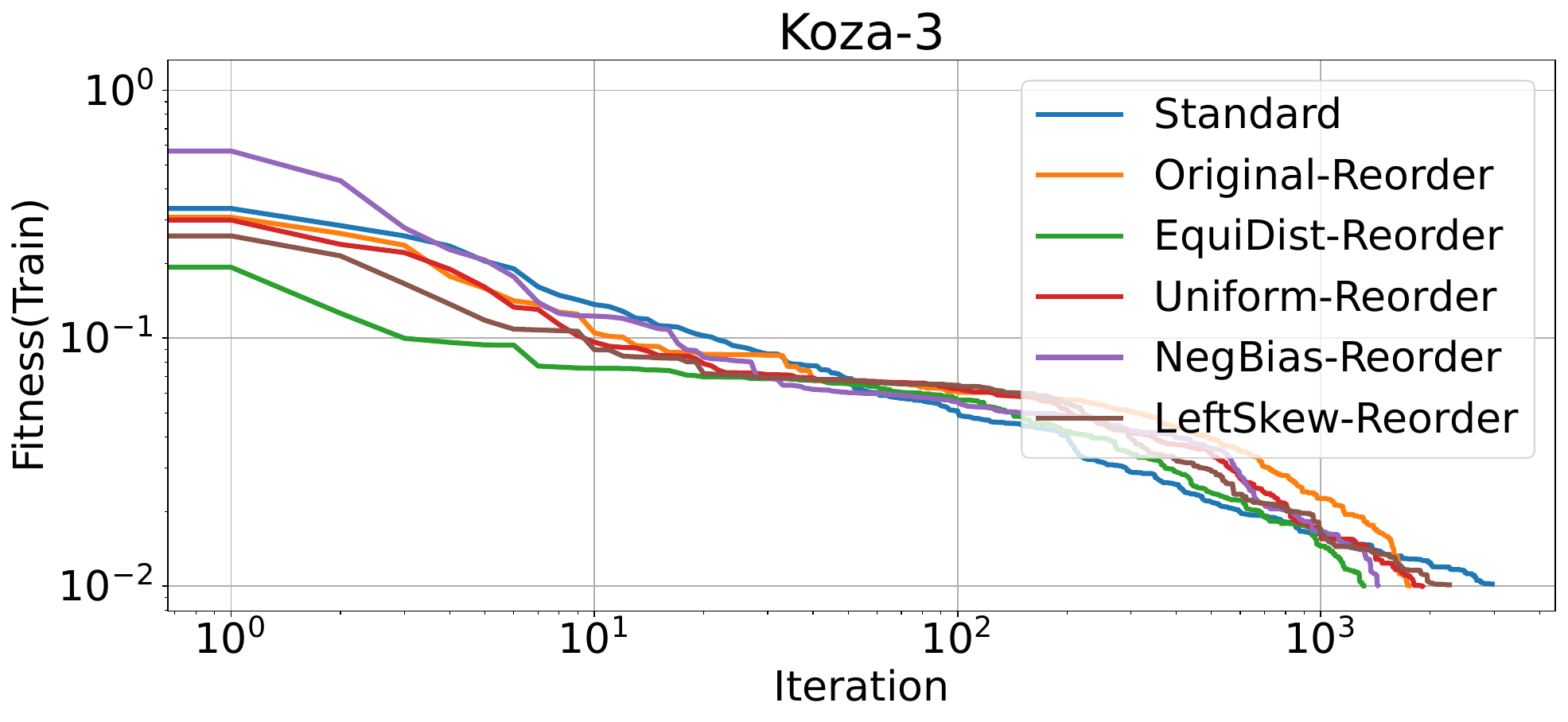}
    
    \includegraphics[width=0.49\textwidth]{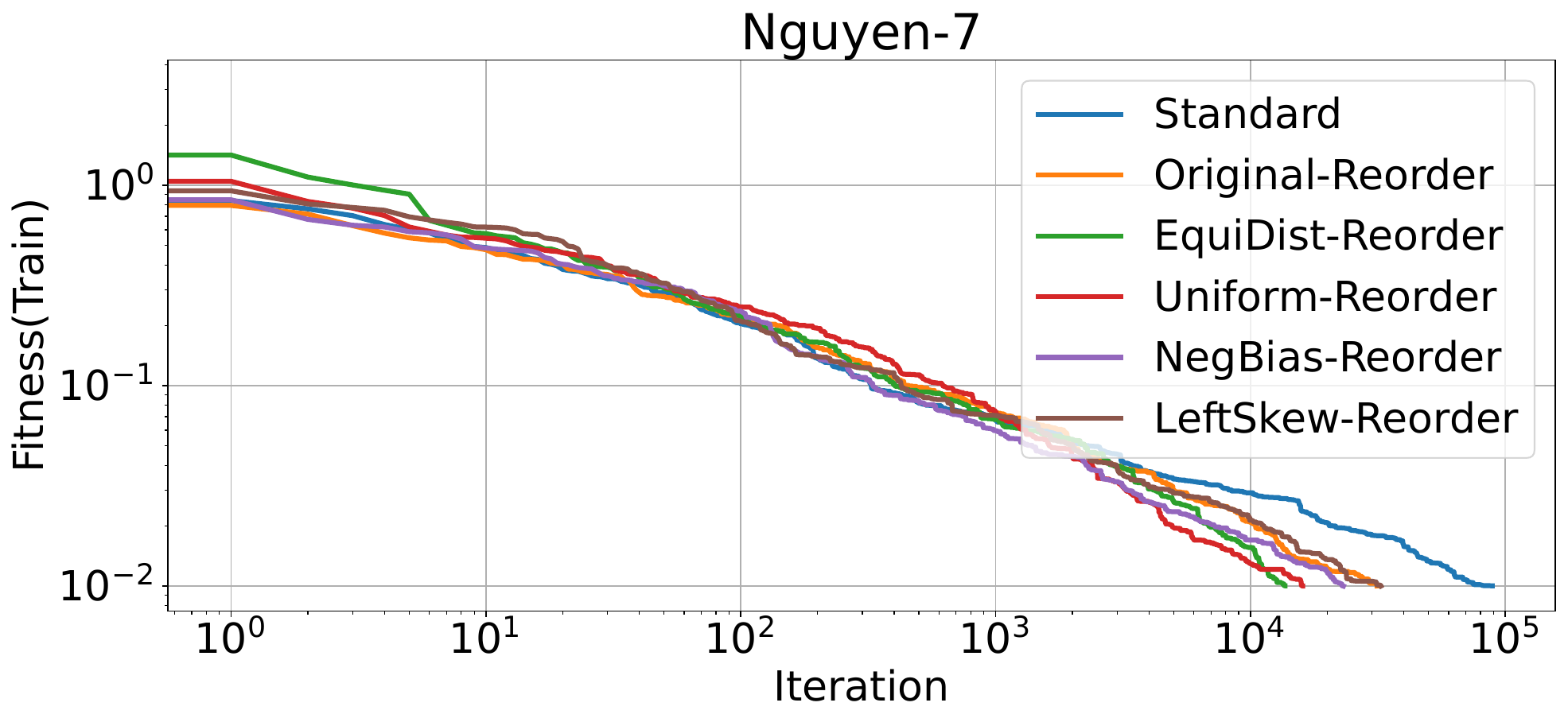}
    \includegraphics[width=0.49\textwidth]{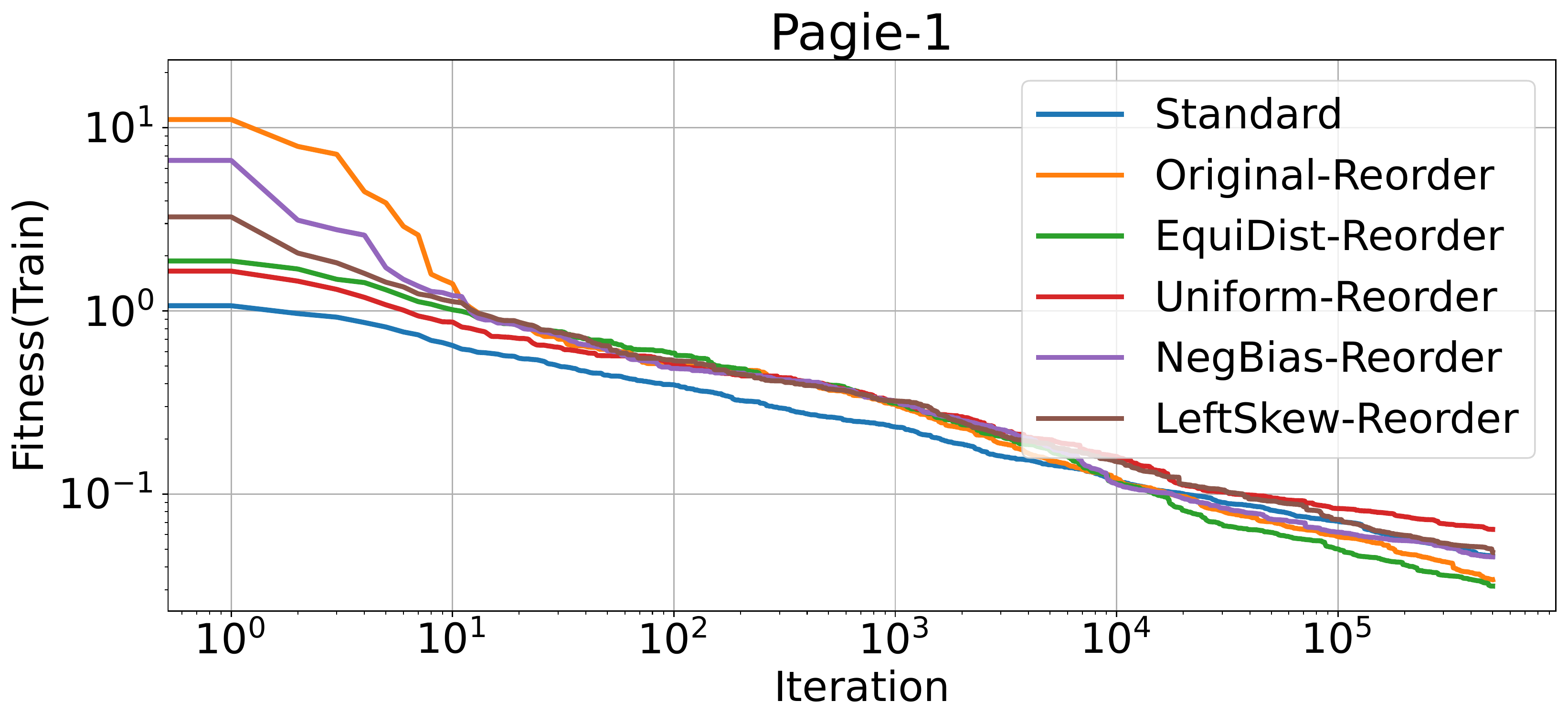}
    
    \caption{Convergence plots for each regression benchmark.
        For better visualization, the x- and y-axis have a logarithmic scale.}
    \label{fig:convergence_plots_regression}
\end{figure}

\begin{figure}[t]%
    \centering
    \includegraphics[width=0.49\textwidth]{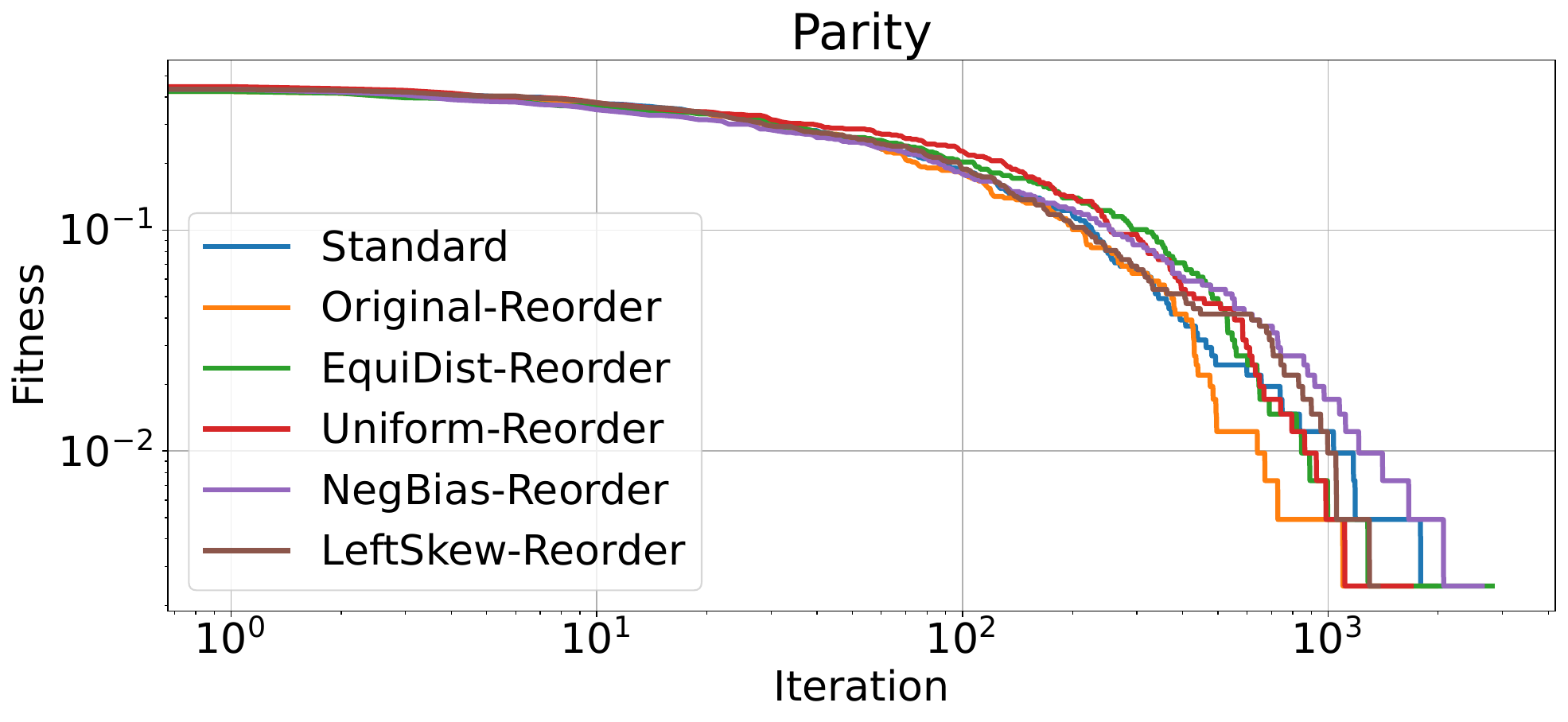}
    \includegraphics[width=0.49\textwidth]{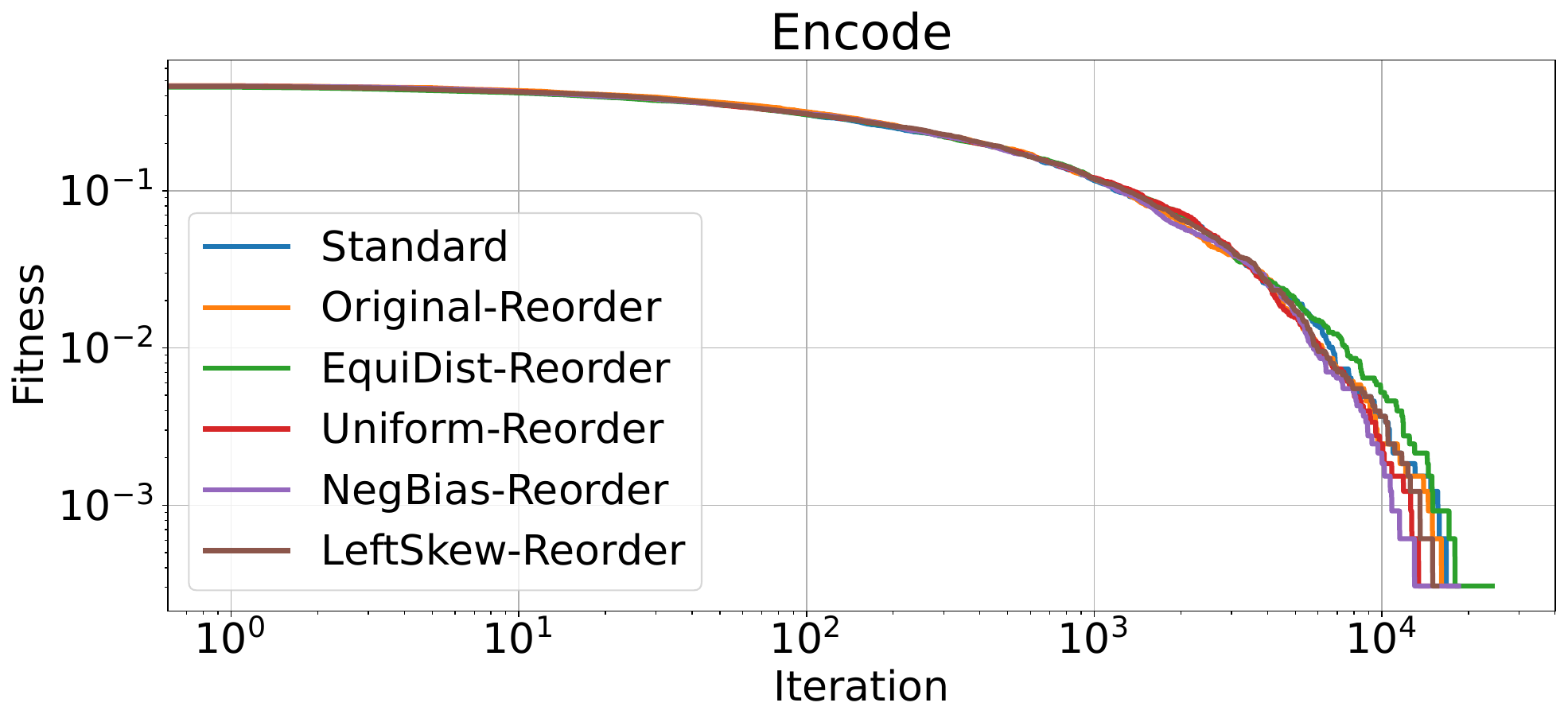}
    
    \includegraphics[width=0.49\textwidth]{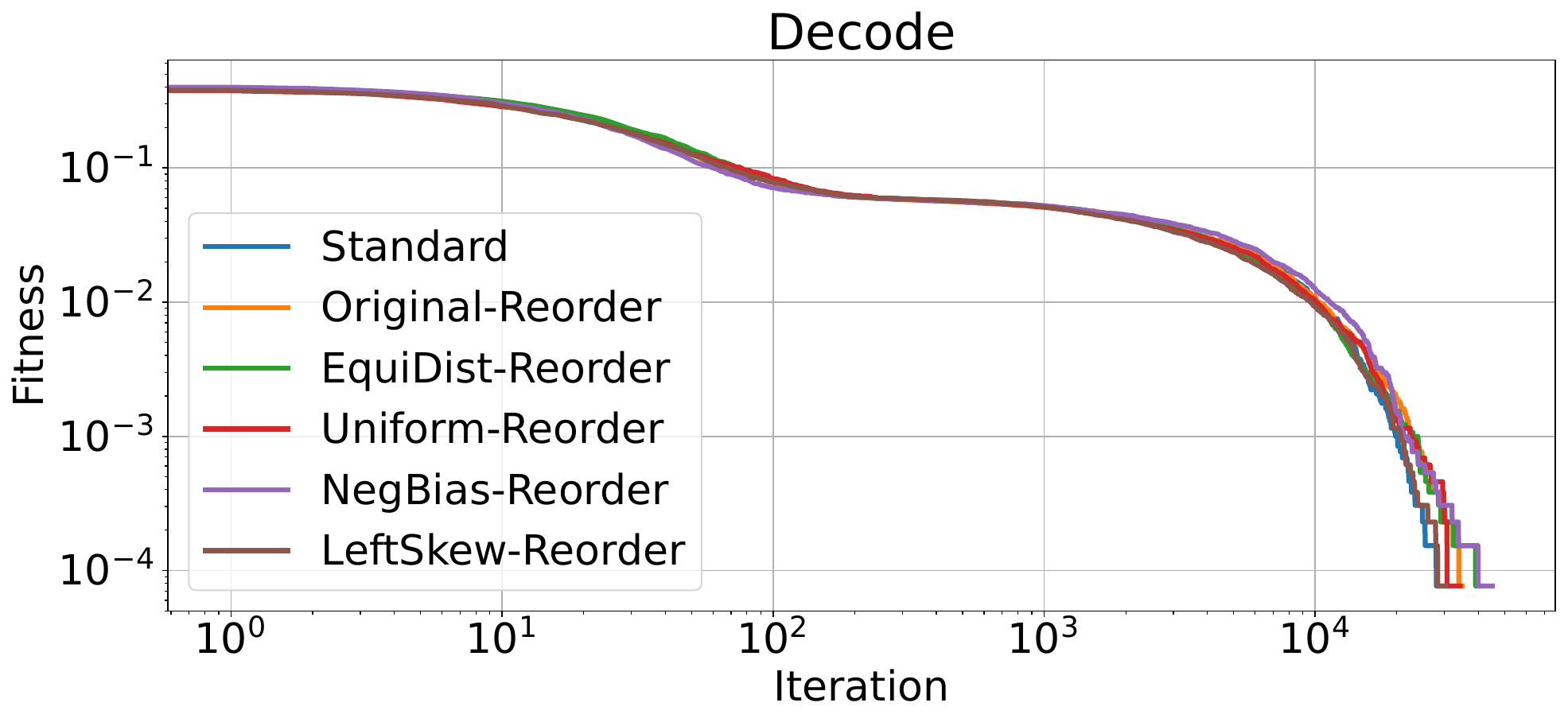}
    \includegraphics[width=0.49\textwidth]{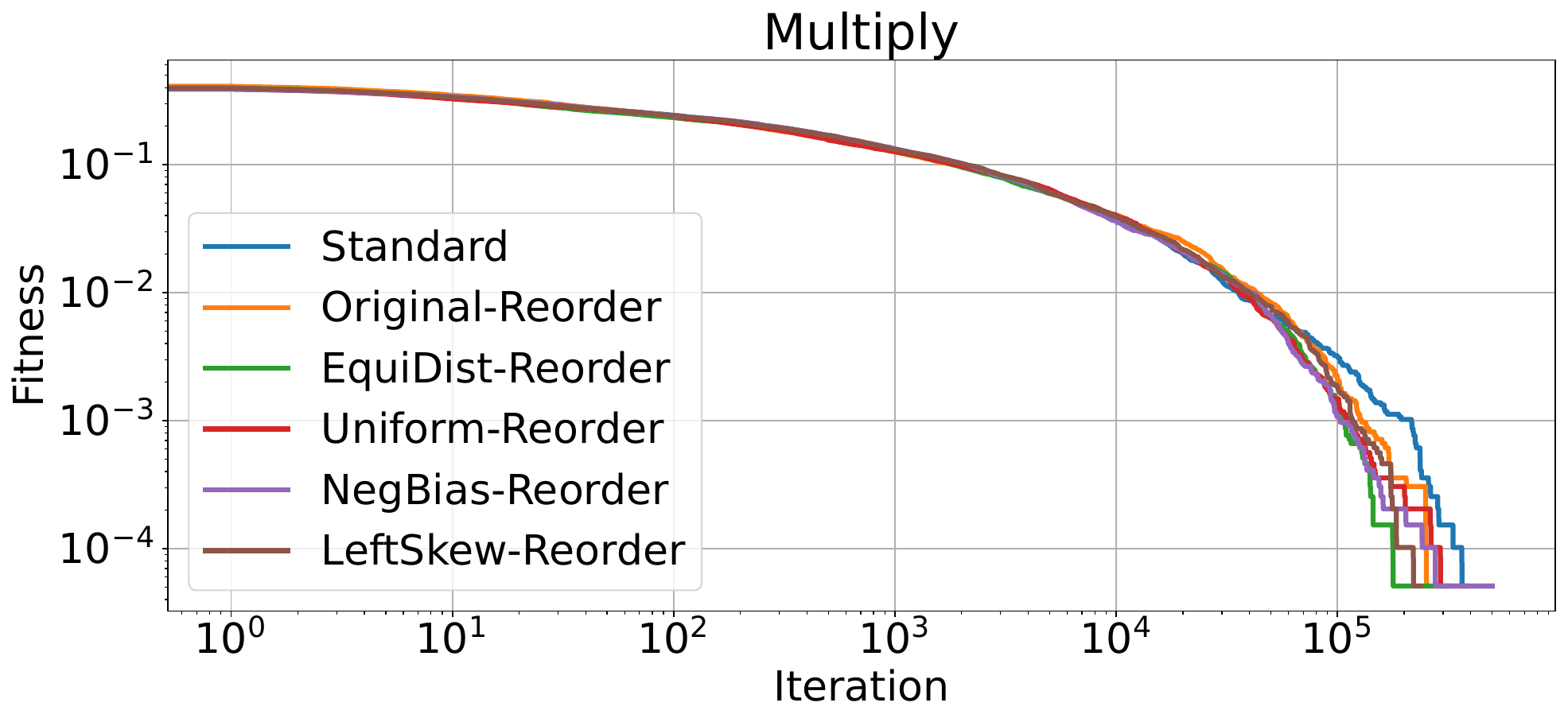}
    
    \caption{Convergence plots for each Boolean benchmark.
        For better visualization, the x- and y-axis have a logarithmic scale.}
    \label{fig:convergence_plots_boolean}
\end{figure}

To better understand the workings of our extensions, we depict convergence plots for both symbolic regression and Boolean benchmarks, as can be seen in Figure~\ref{fig:convergence_plots_regression} and Figure~\ref{fig:convergence_plots_boolean} respectively.
For these plots, we averaged the convergence of 75 runs.
Furthermore, to easier see the differences, the x- and y-axis have a logarithmic scale.
For unscaled plots, we refer to Figure~\ref{fig_appendix:convergence_plots} in Appendix~\ref{app:convergence_graphs}.
In addition, their respective standard deviation is included in Figure~\ref{fig_appendix:convergence_plots_with_std}.

When the plots for symbolic regression benchmarks are examined, they all show very similar behaviour.
According to the categorization of Stegherr et al.~\cite{Stegherr23}, their convergence behaviour can all be grouped into the category \emph{Fast to Slow}.
Within the first few iterations, a relatively low fitness value is achieved.
Afterwards, the rate of improvement decreases for all CGP variants, and 
a lot of training iterations are needed for small improvements.
Interestingly, \EReorder{} and \UDReorder{}, as well as \NPBReorder{} and \LSDReorder{}, respectively, did not show similar performances as can be seen in Section~\ref{subsec:performance_boolean} and Section~\ref{subsec:performance_regression}.
Nevertheless, when only convergence plots are examined, these reorder operators behave very similar.
Thus, it can be concluded that the behaviour of CGP does not change when a shuffling method is included.

CGP's behaviour in accordance to Boolean benchmarks are all identical, as is depicted in Figure~\ref{fig:convergence_plots_boolean}.
Again, their behaviour can be categorized into \emph{Fast to Slow}~\cite{Stegherr23}.
As all CGP variants behave the same, no behavioural change can be concluded again when a reorder method is included or not.
However, please note that Boolean benchmarks show a \emph{deceptive} fitness landscape~\cite{Vasicek2018}.
As a multitude of different solutions lead to the same fitness value, examining the convergence plot might not lead to reasonable findings or might even be deceptive.
This is why a conclusion based on Boolean benchmarks should be met with scepticism.

\section{Conclusion}
\label{sec:conclusion}

In this work, we \emph{extended reorder methods} used by Cartesian Genetic Programming (CGP).
We expanded upon the \emph{original reorder} method (\OGReorder{}) from Goldman and Punch~\cite{dag_origin}, as well as another reorder version called \emph{Equidistant Reorder} (\EReorder{}) from Cui et al.~\cite{cui_ecta_reorder}.
Three novel operators were introduced which shuffle CGP's genotype and assigns active nodes new positions in the genotype.
Just as the two existing operators, the ordering of active nodes is preserved for all extensions, which means that the phenotype does not change.
One of our new operators is called \emph{Uniformly Distributed Reorder} (\UDReorder{}) and builds upon \EReorder{}.
Instead of placing active nodes equidistantly apart---which is done by \EReorder{}---their new positions are sampled from a continuous uniform distribution.
The second novel extension is called \emph{Negative Positional Bias based Reorder} (\NPBReorder{}).
It moves all active nodes to the end of the genotype, just before output nodes.
The last operator, called \emph{Left-Skewed Distribution Reorder} (\LSDReorder{}), is inspired by \NPBReorder{}.
Instead of using fixed placements, the positioning of active nodes is determined by a left-skewed beta distribution.

To gauge the effectiveness of the new operators, we conducted an empirical study.
A total of six algorithms were compared: the standard CGP formula (\Standard{}), \OGReorder{}, \EReorder{}, \UDReorder{}, \NPBReorder{} and \LSDReorder{}.
We tested them on four Boolean- and four symbolic regression benchmarks, and optimized the hyperparameters for each setting.
To find the best hyperparameters and to rank the effectiveness of algorithms, \emph{Bayesian data analysis} for the posterior distributions of results was performed.

Our findings show that, in all cases, CGP including a reorder operator is able to surpass \Standard{} regarding its \emph{number of training iterations until a solution is found (I2S)} and/or fitness value.
However, there is no reorder algorithm that is always able to deliver the best results for all benchmarks and the choice of the right shuffling algorithm highly depends on the benchmark.
Furthermore, while \NPBReorder{} or \LSDReorder{} may be the best choice for some benchmarks, they depend on an additional hyperparameter.
Thus, the complexity to find the best hyperparameters increases which leads to a trade-off between better performance but higher computational overhead beforehand.
Regarding symbolic regression benchmarks, some reorder methods might lead to a better test fitness value but also increase the I2S.
This, again, might lead to a trade-off in some specific cases.
Considering harder problems, \EReorder{} always outperforms \UDReorder{}. 
This indicates that enforcing an equidistant spacing should not limit CGP.

Algorithmically, \EReorder{} and \UDReorder{}, as well as \NPBReorder{} and \LSDReorder{}, are close.
\EReorder{} enforces an equidistant spacing, while \UDReorder{} samples from a continuous uniform distribution.
\NPBReorder{} moves all active nodes near output nodes, while \LSDReorder{} samples from a beta distribution to approximate \NPBReorder{}.
However, their respective fitness values greatly differ sometimes. 

Additionally, we investigated their respective convergence behaviour.
Interestingly, \emph{all} CGP versions show a \emph{Fast to Slow} convergence plot~\cite{Stegherr23}.
However, the behaviour for Boolean benchmarks must be met with reservation because of their \emph{deceptive} fitness landscape~\cite{Vasicek2018}.

For future works, their behaviour can be further analysed.
While the convergence plots of \EReorder{} and \UDReorder{}---as well as \NPBReorder{} and \LSDReorder{}---depict the same behaviour, their fitness value might differ greatly. 
As they are algorithmically close, the reason for their gap in performance is still an open question.
Answering this question might even reveal other hidden biases or properties of CGP.

\bmhead{Acknowledgments}
The authors would like to thank the German Federal Ministry of Education and Research (BMBF) for supporting the project SaMoA within VIP+ (grant number 03VP09291).

\backmatter

\section*{Declarations}

\section*{Funding}
Partial financial support for this work was received from the German Federal Ministry of Education and Research (BMBF) via the project SaMoA within VIP+ (grant number 03VP09291).

\section*{Conflict of interest/Competing interests}
Not applicable

\section*{Ethics approval}
Not applicable

\section*{Consent to participate}
Not applicable

\section*{Consent for publication}
Not applicable

\section*{Availability of data and materials}
Not applicable

\section*{Code availability}
The source code including benchmarks used to produce the experimental data is available at: \url{https://github.com/CuiHen/Reorder_Strategies_for_CGP}

\section*{Authors' contributions}
Henning Cui developed the operators, designed and performed experiments, analysed the data and wrote the manuscript.
Andreas Margraf contributed to the conceptualization of the new operators.  
J\"org H\"ahner supervised the project and also contributed to the conceptualization of the new operators.
All authors discussed the results and commented on the manuscript at all stages.

%
%
%
%

\begin{appendices}
    
    \section{Convergence Plots Without Altered Axis}
    \label{app:convergence_graphs}
    
    \begin{figure}[t]%
        \centering
        \includegraphics[width=0.49\textwidth]{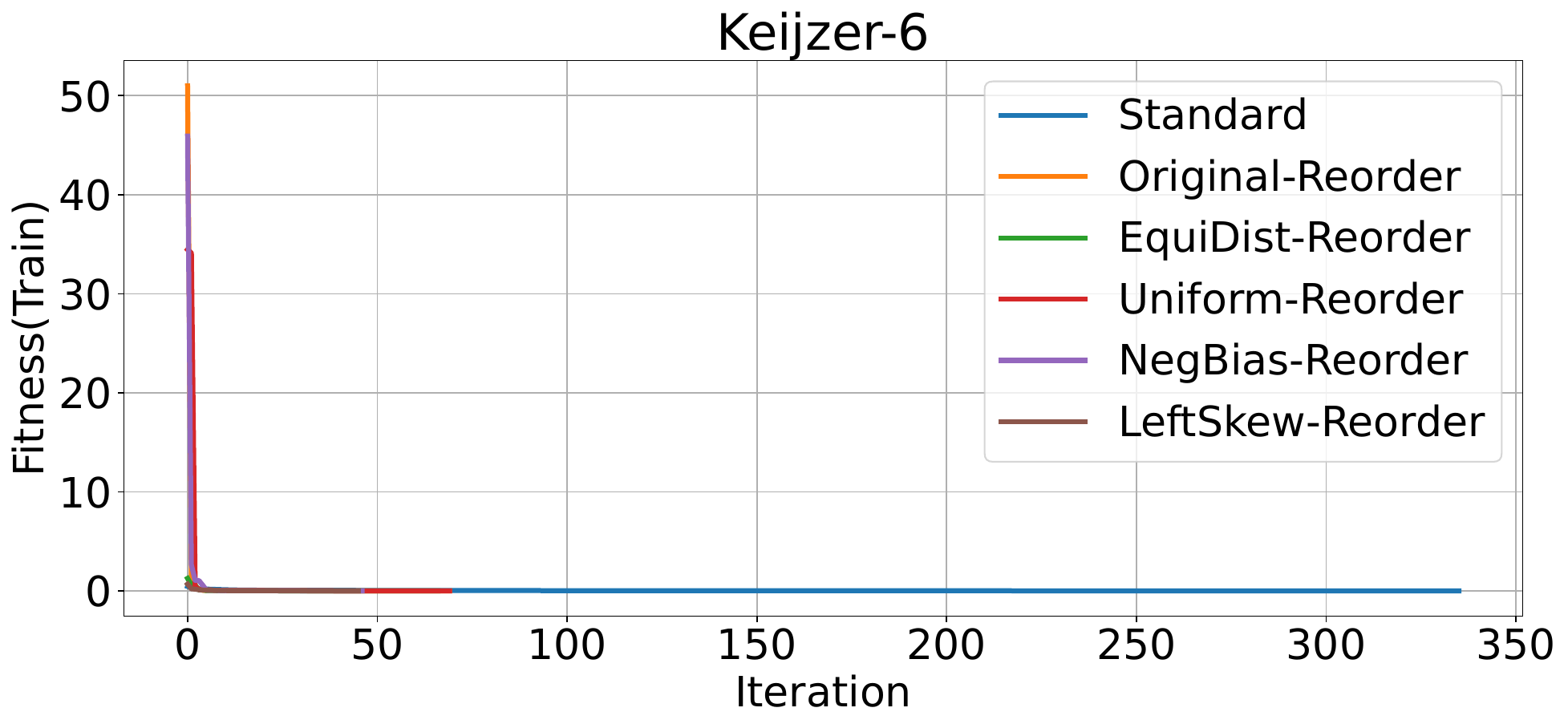}
        \includegraphics[width=0.49\textwidth]{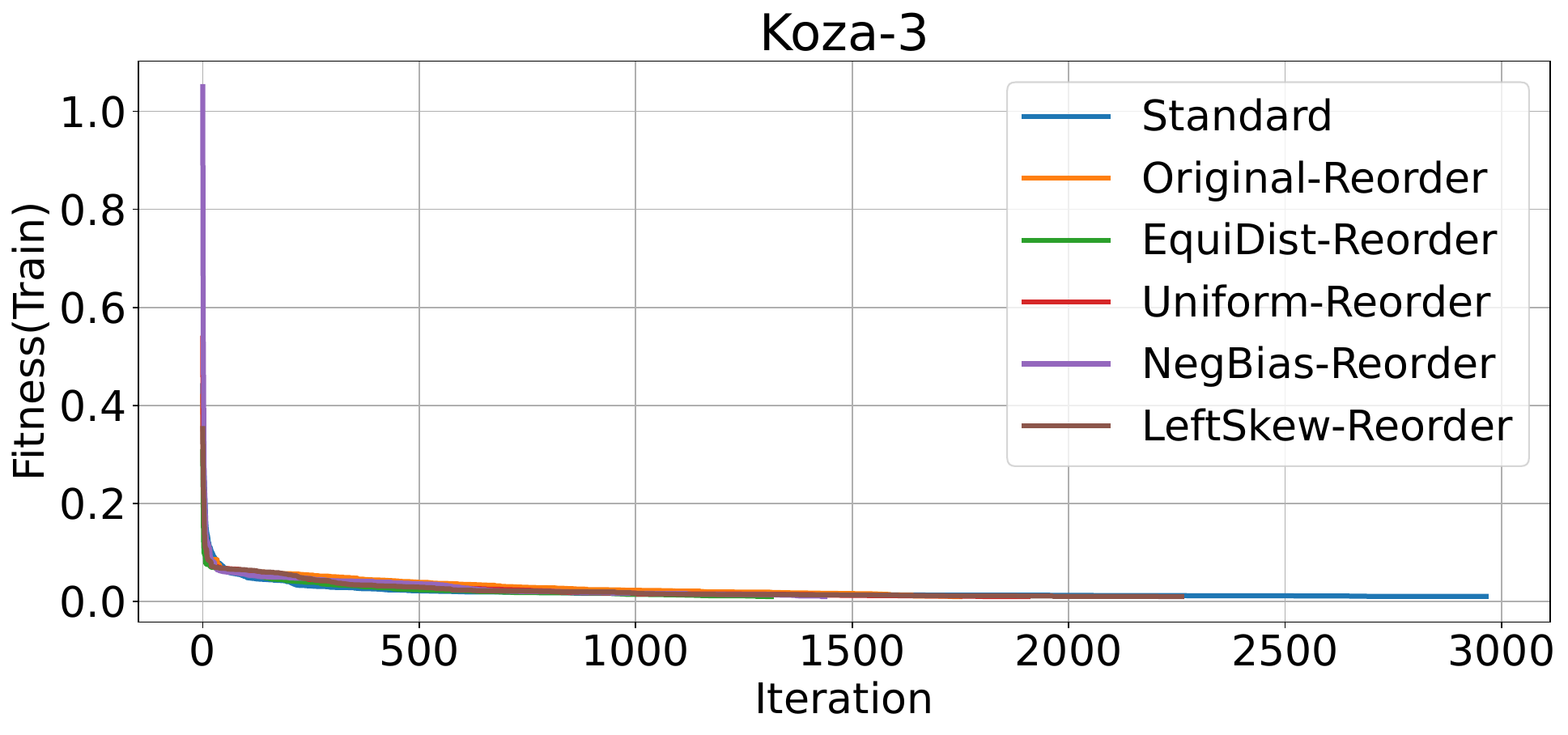}
        
        \includegraphics[width=0.49\textwidth]{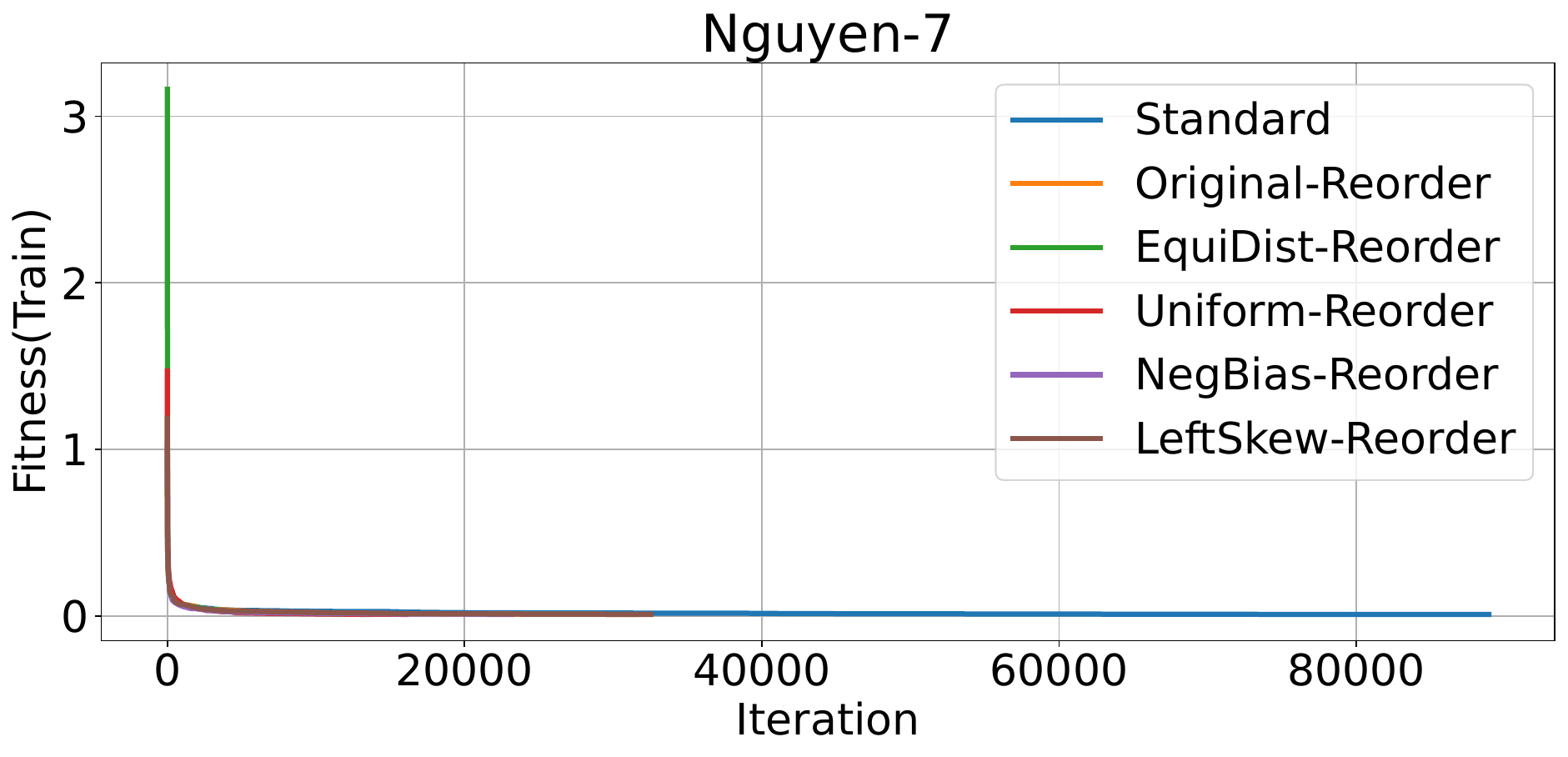}
        \includegraphics[width=0.49\textwidth]{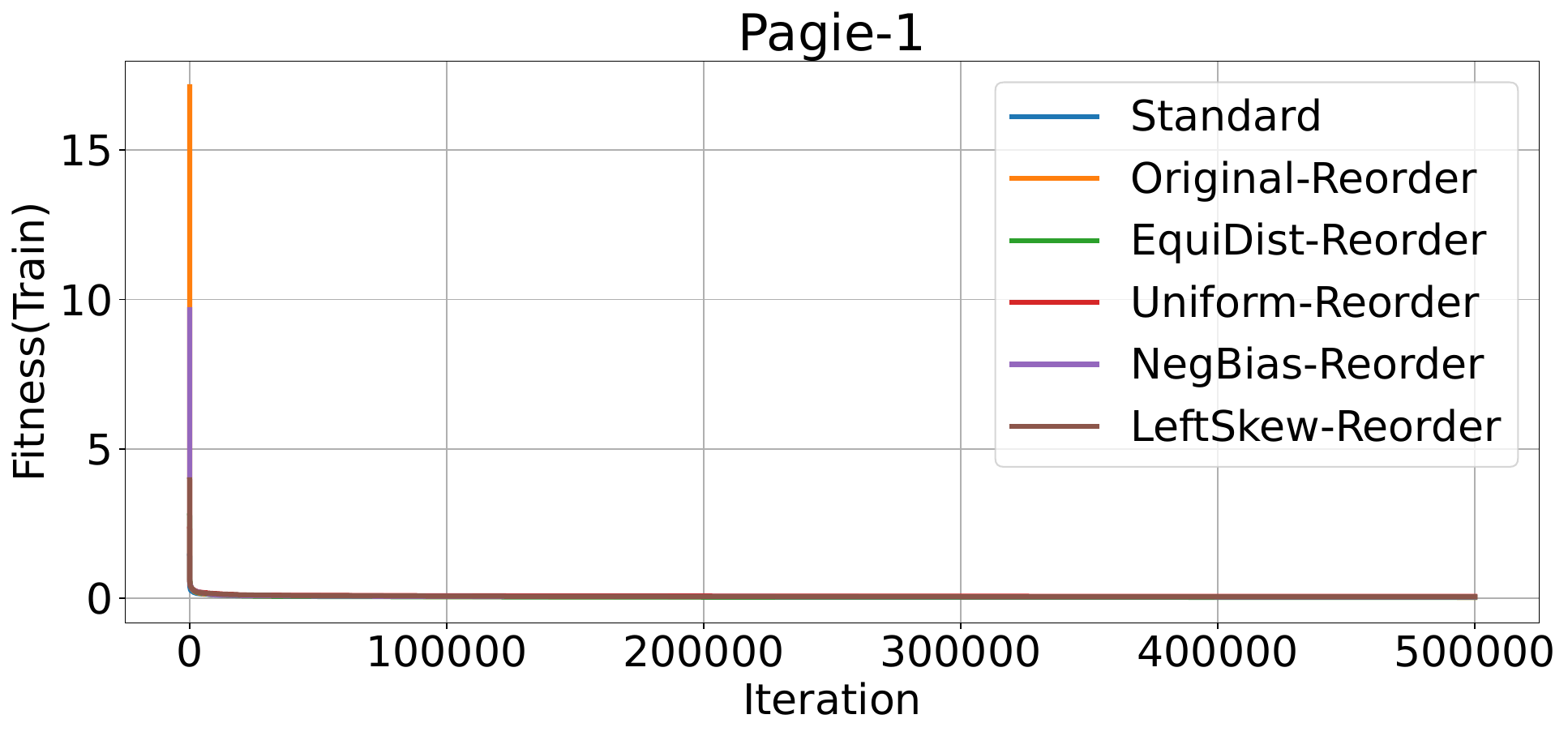}
        
        \includegraphics[width=0.49\textwidth]{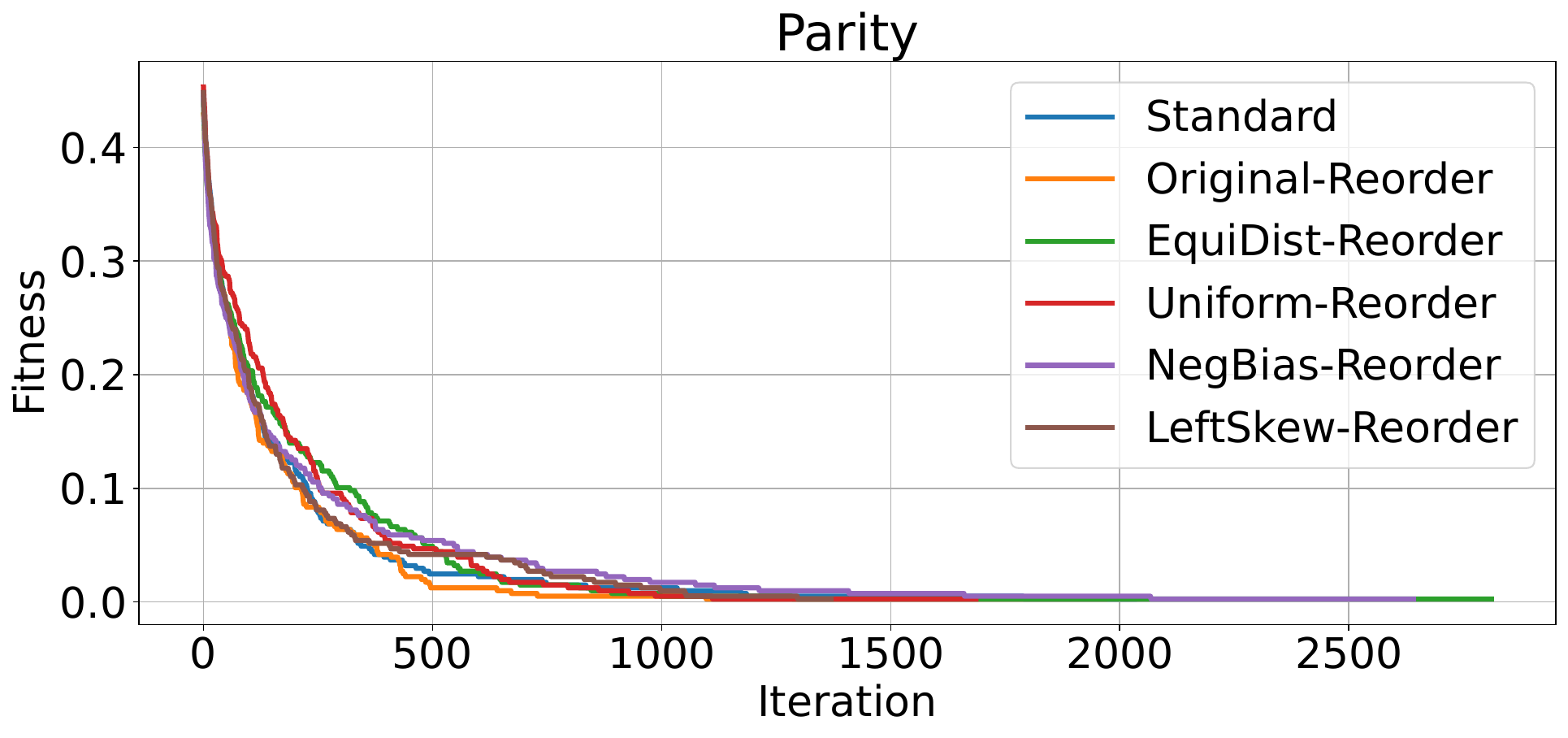}
        \includegraphics[width=0.49\textwidth]{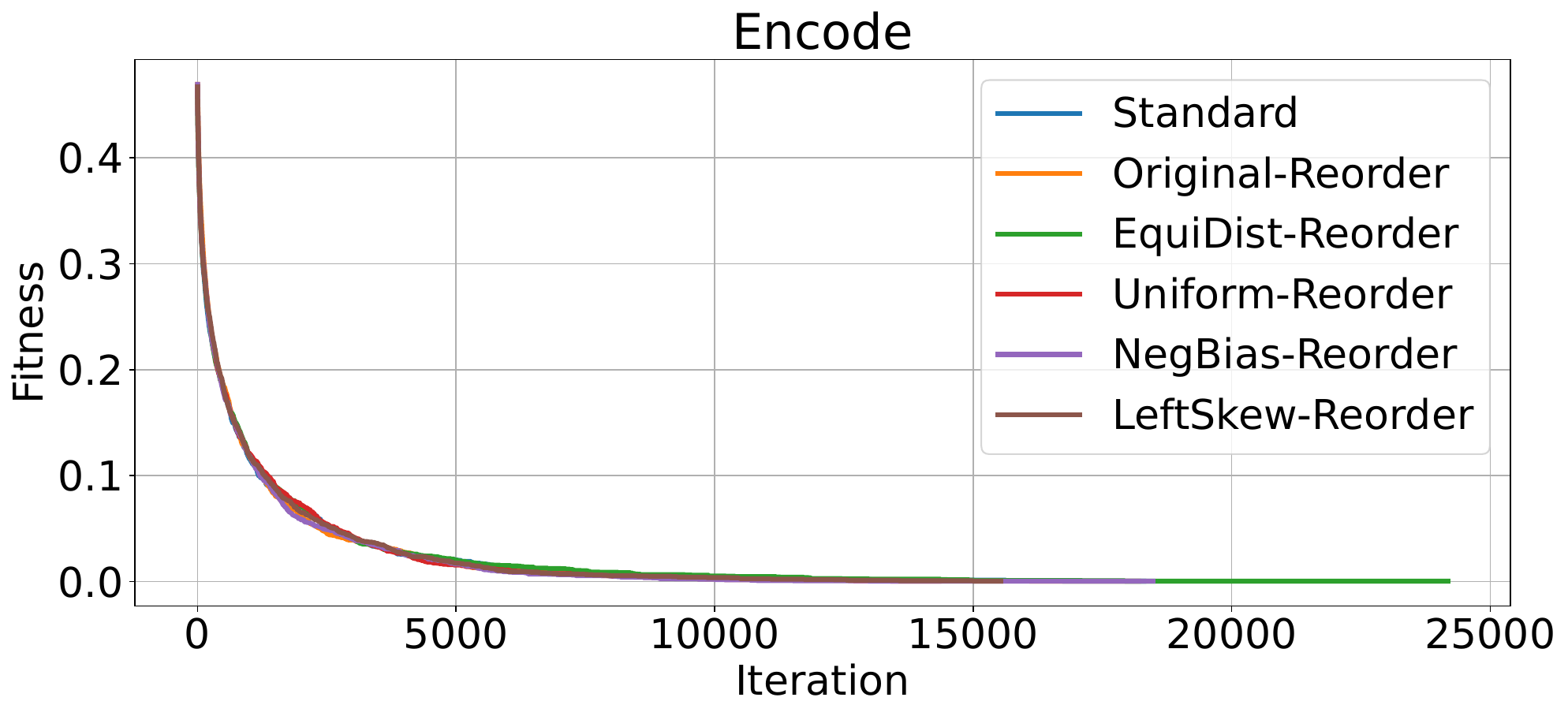}
        
        \includegraphics[width=0.49\textwidth]{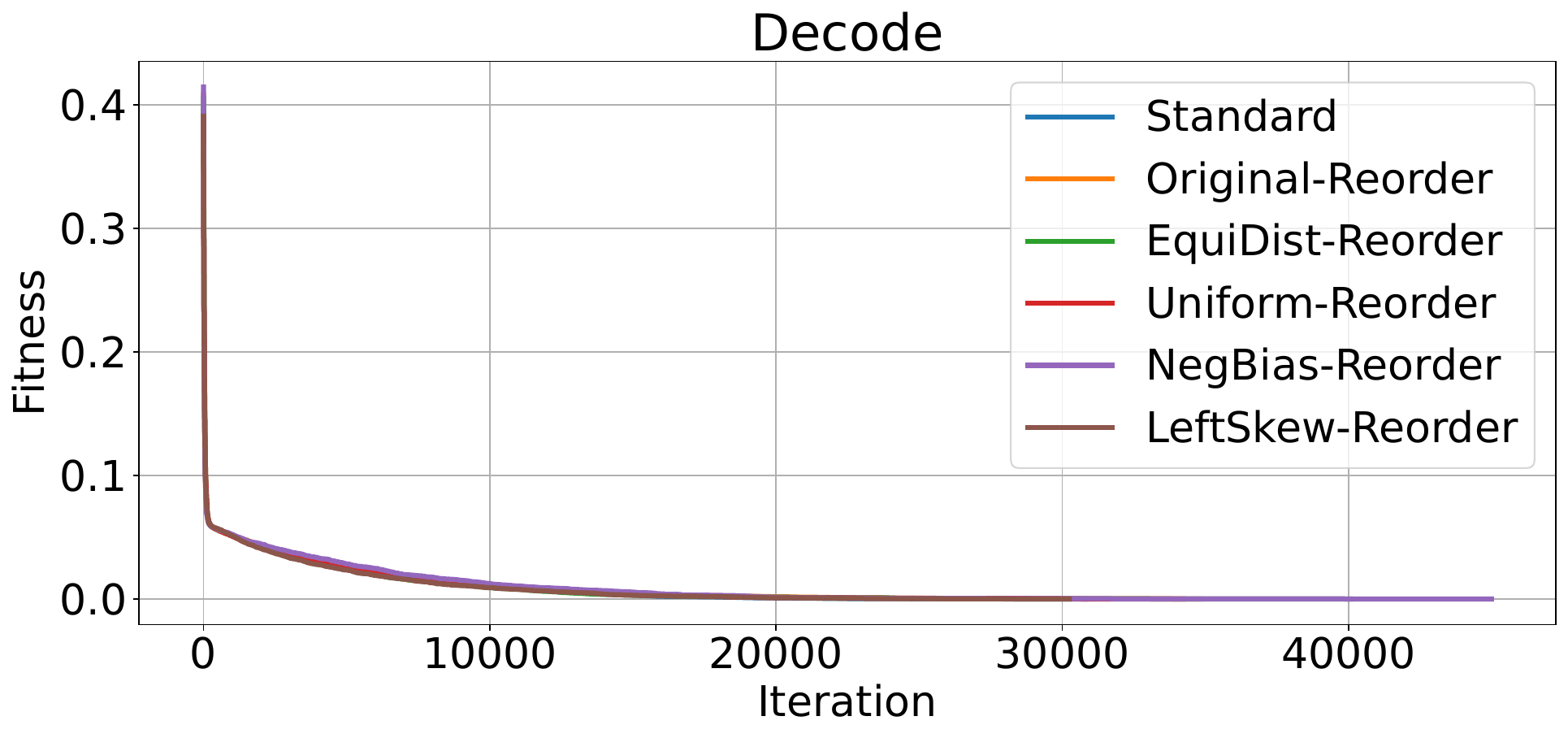}
        \includegraphics[width=0.49\textwidth]{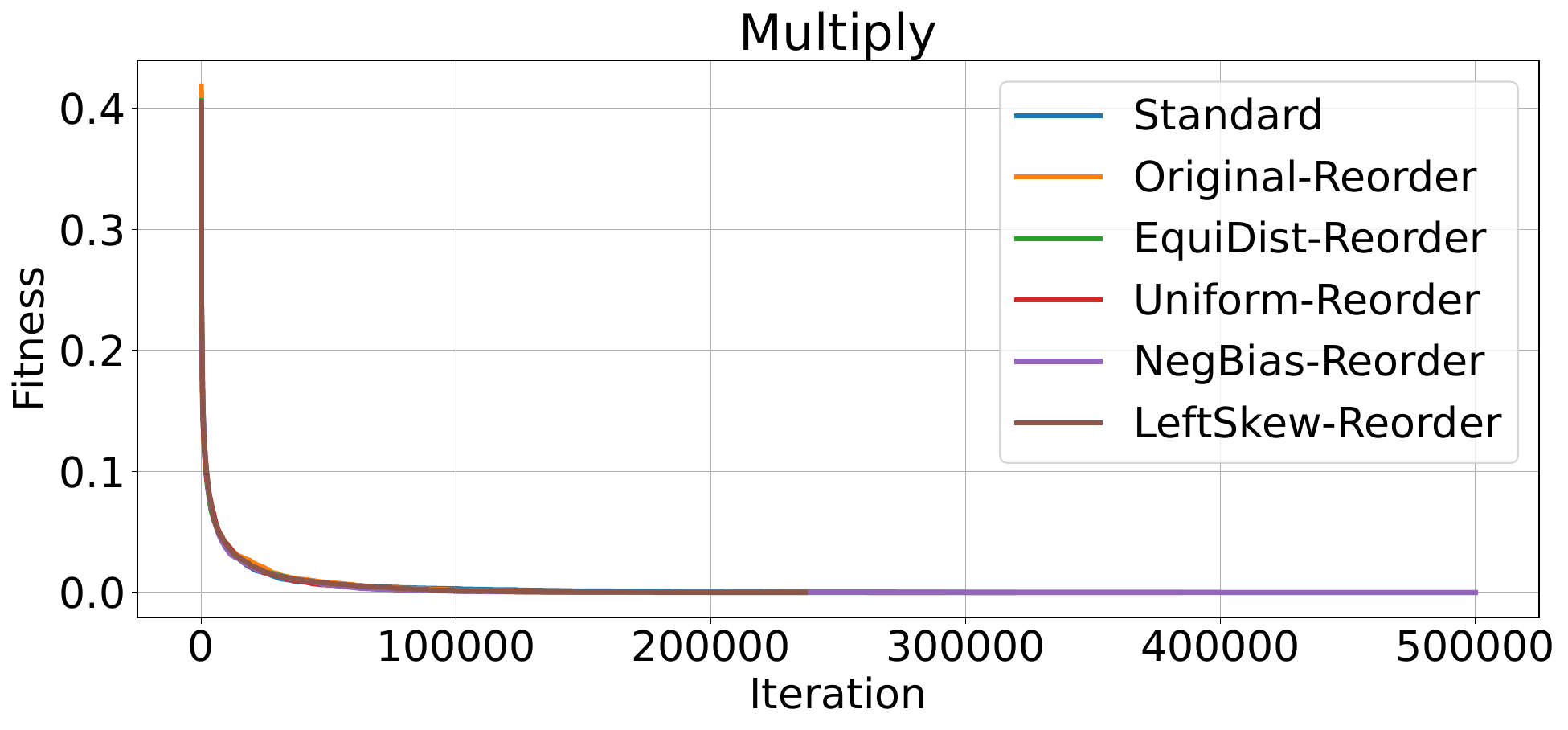}
        
        \caption{Convergence plots for each symbolic regression and Boolean benchmark. Scales have not been altered.}
        \label{fig_appendix:convergence_plots}
    \end{figure}
    
    \begin{figure}[t]%
        \centering
        \includegraphics[width=0.49\textwidth]{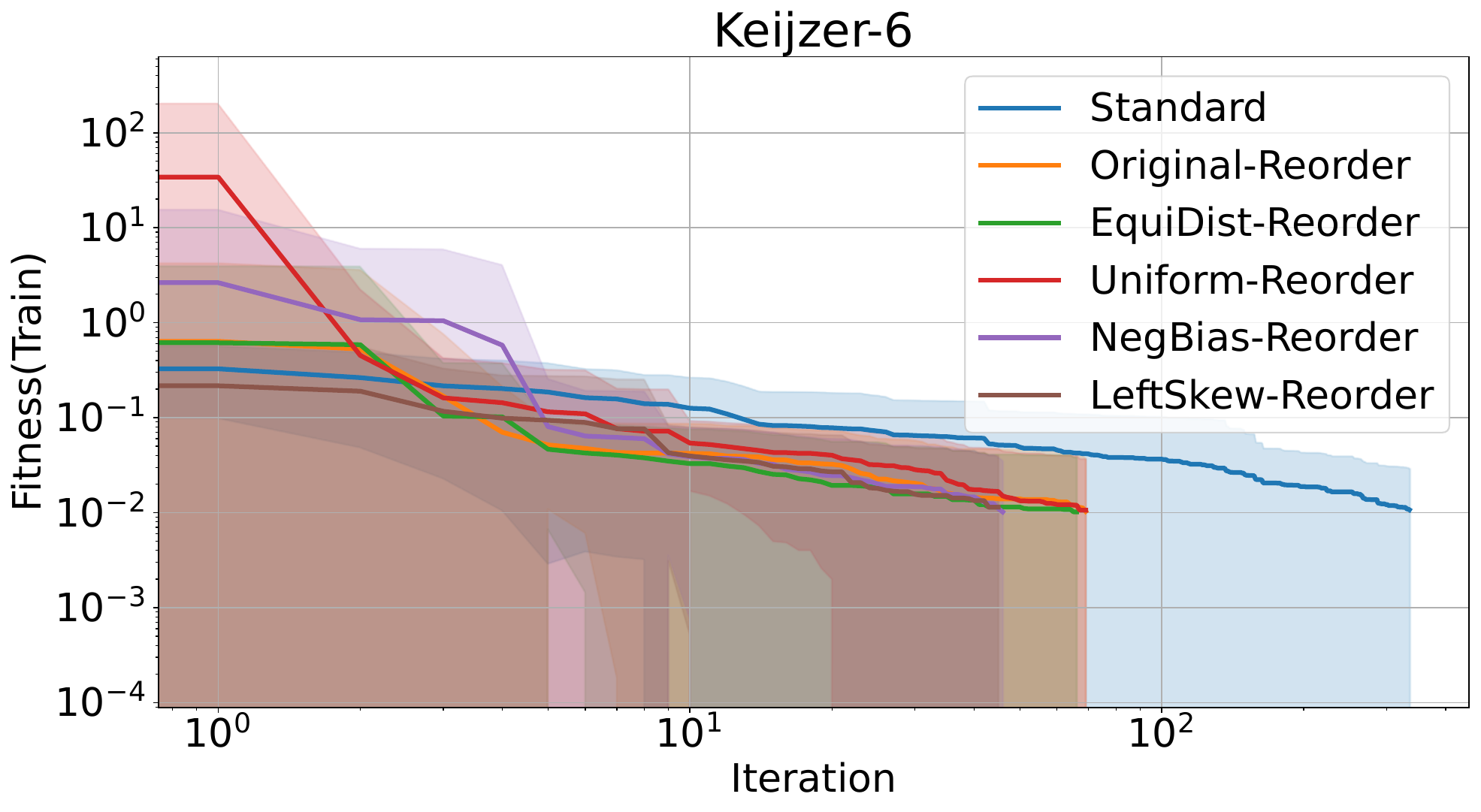}
        \includegraphics[width=0.49\textwidth]{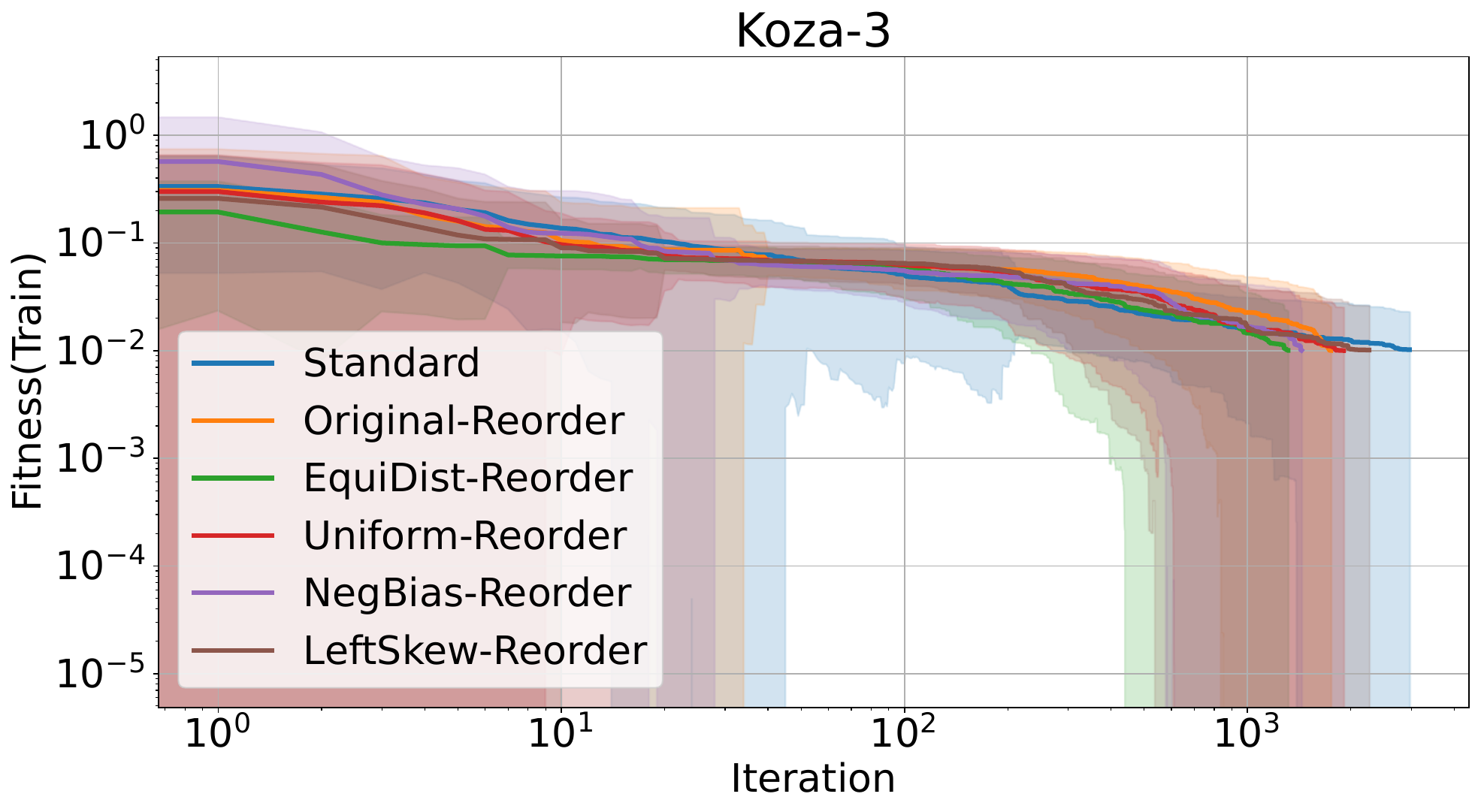}
        
        \includegraphics[width=0.49\textwidth]{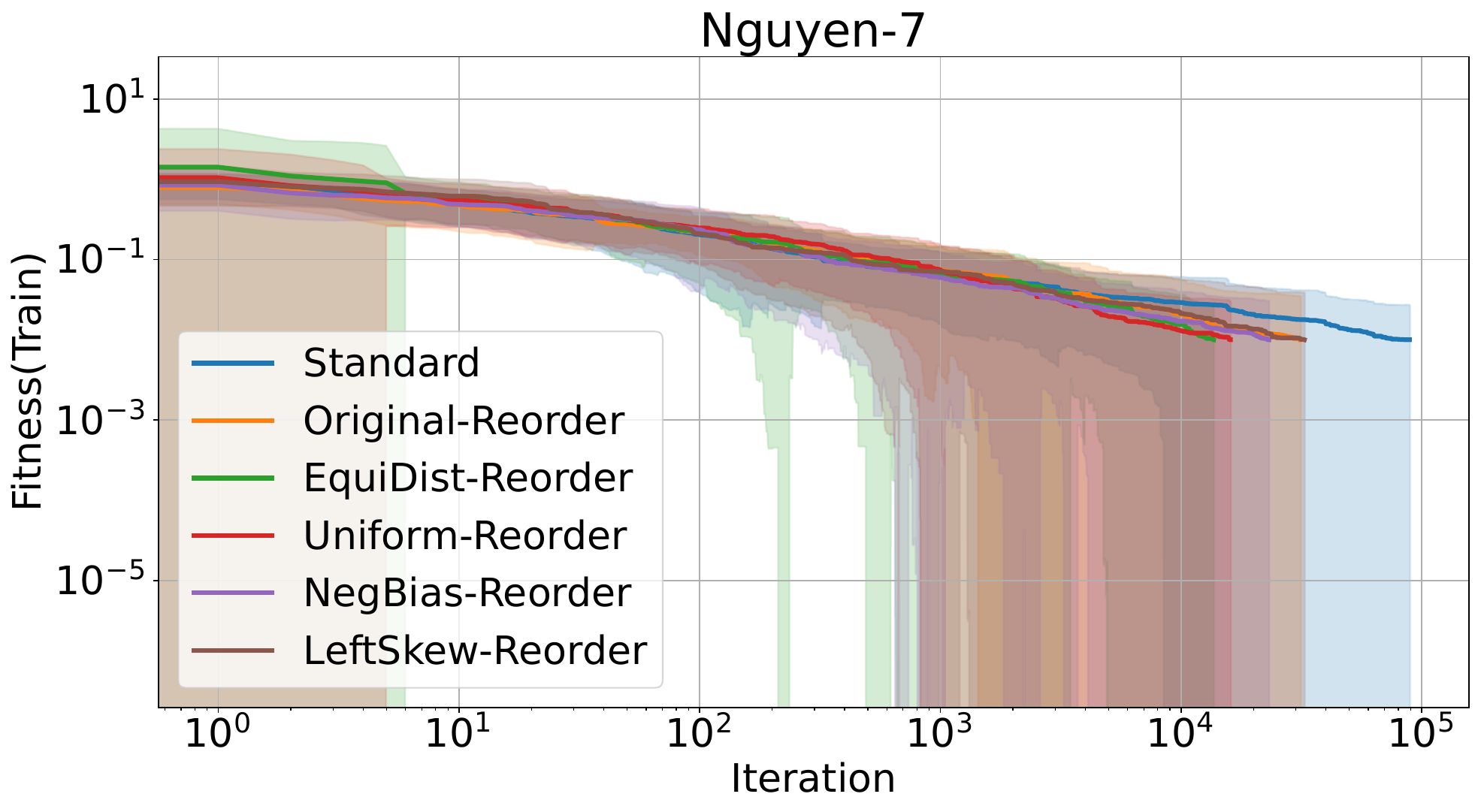}
        \includegraphics[width=0.49\textwidth]{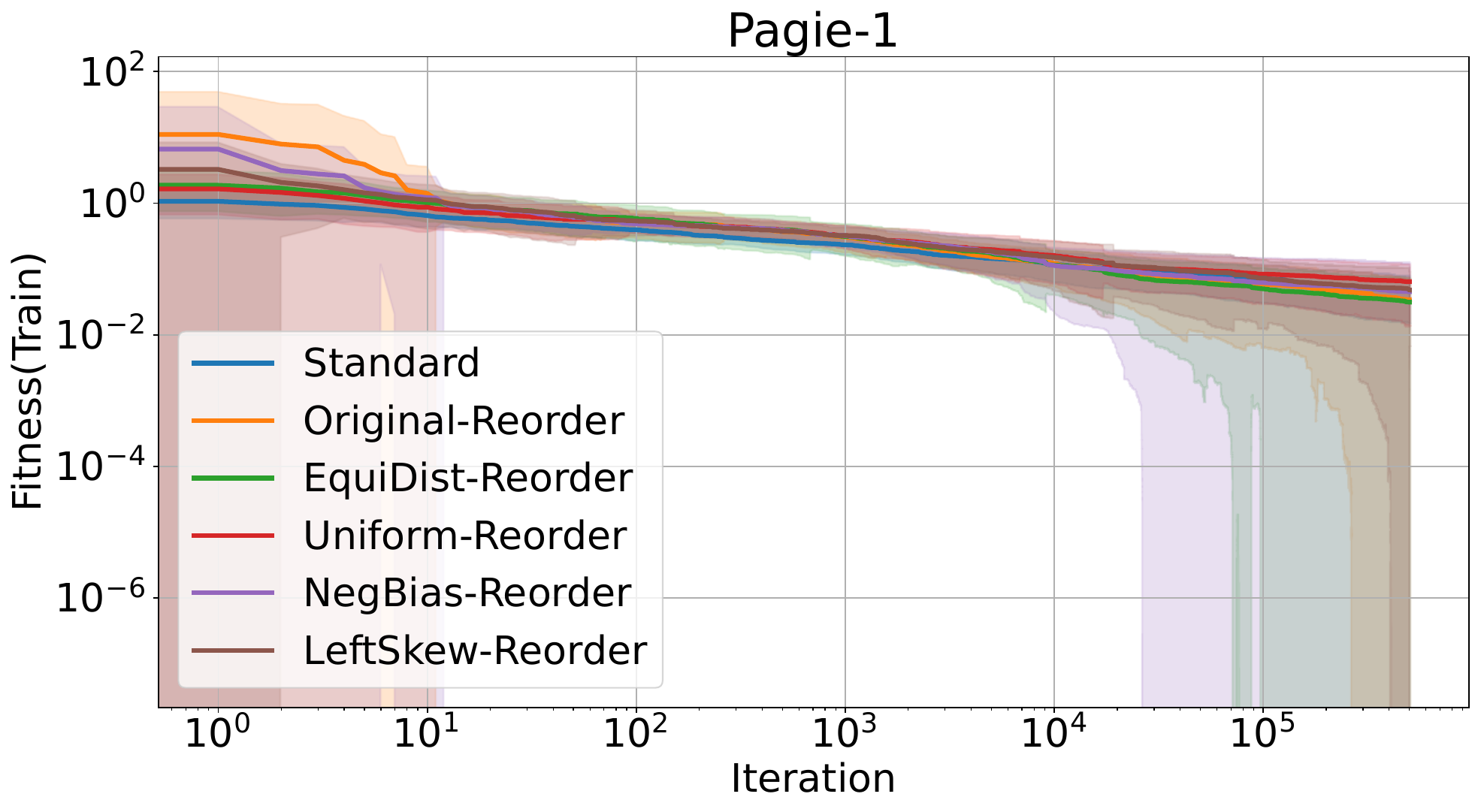}
        
        \includegraphics[width=0.49\textwidth]{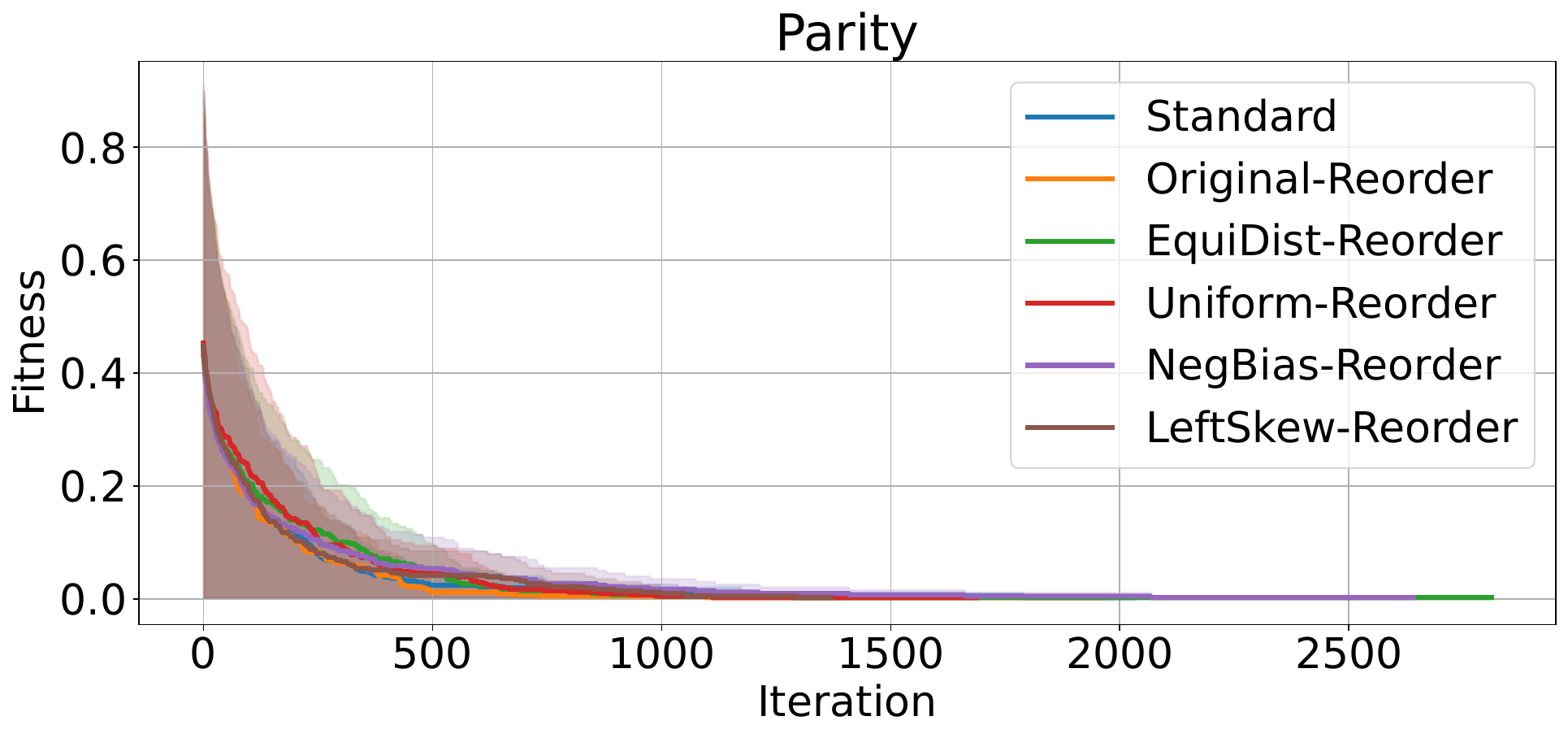}
        \includegraphics[width=0.49\textwidth]{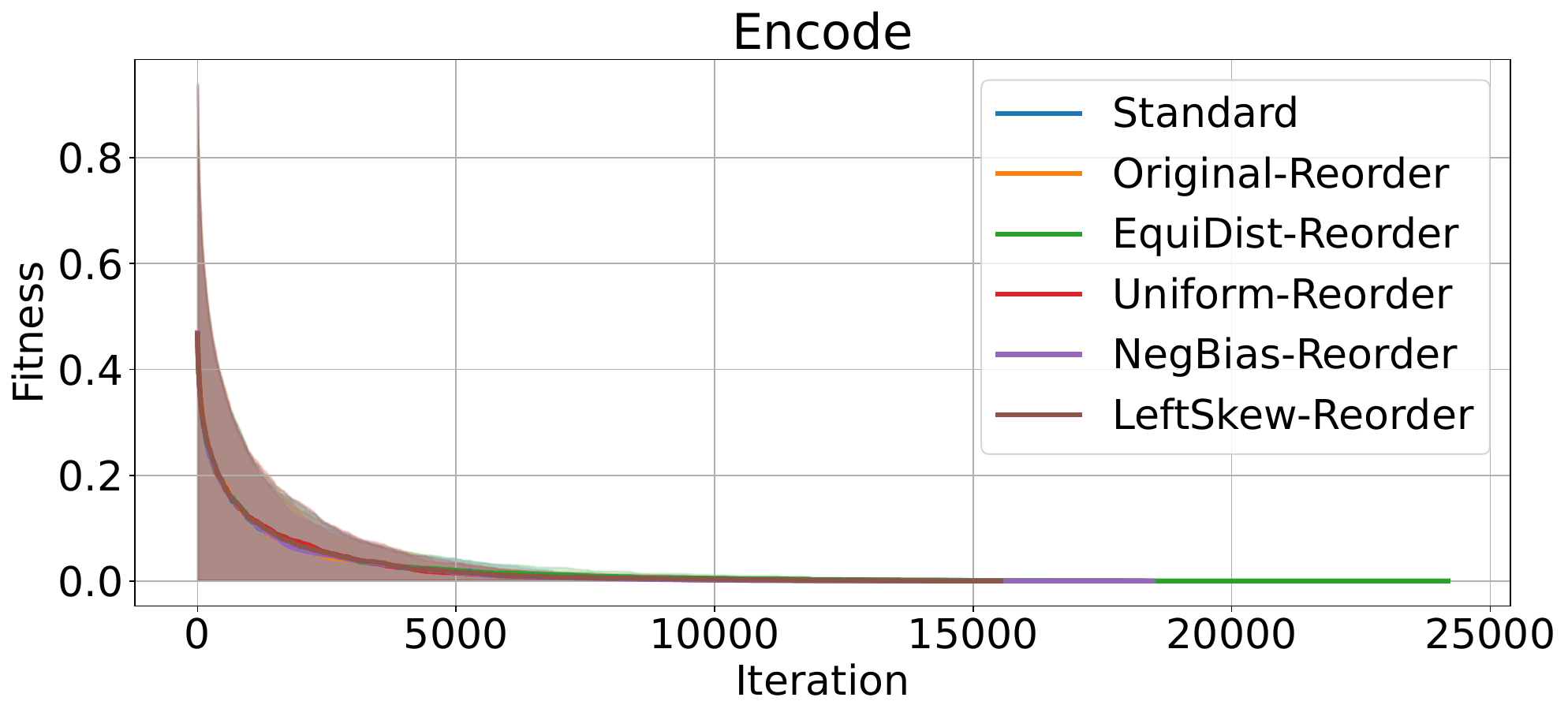}
        
        \includegraphics[width=0.49\textwidth]{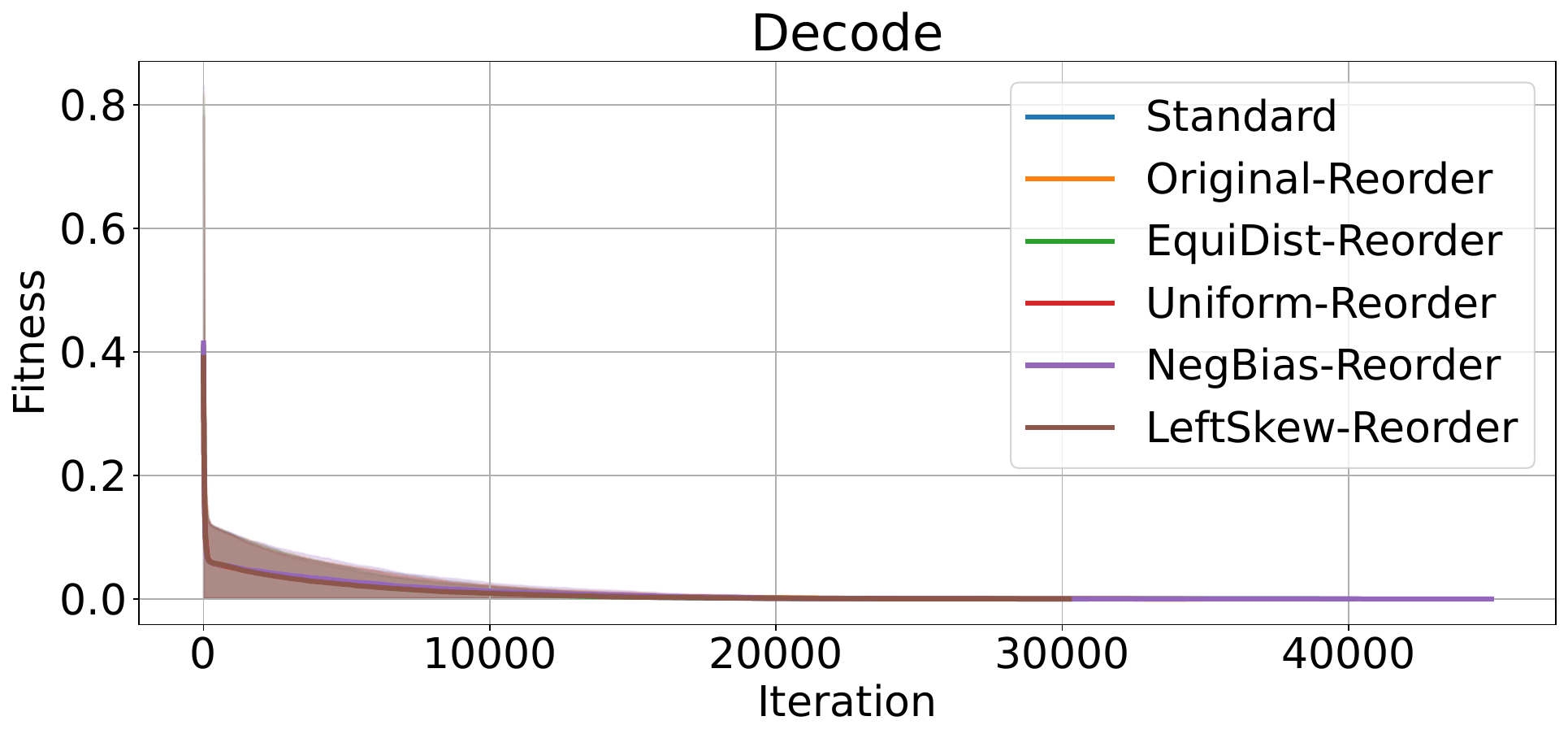}
        \includegraphics[width=0.49\textwidth]{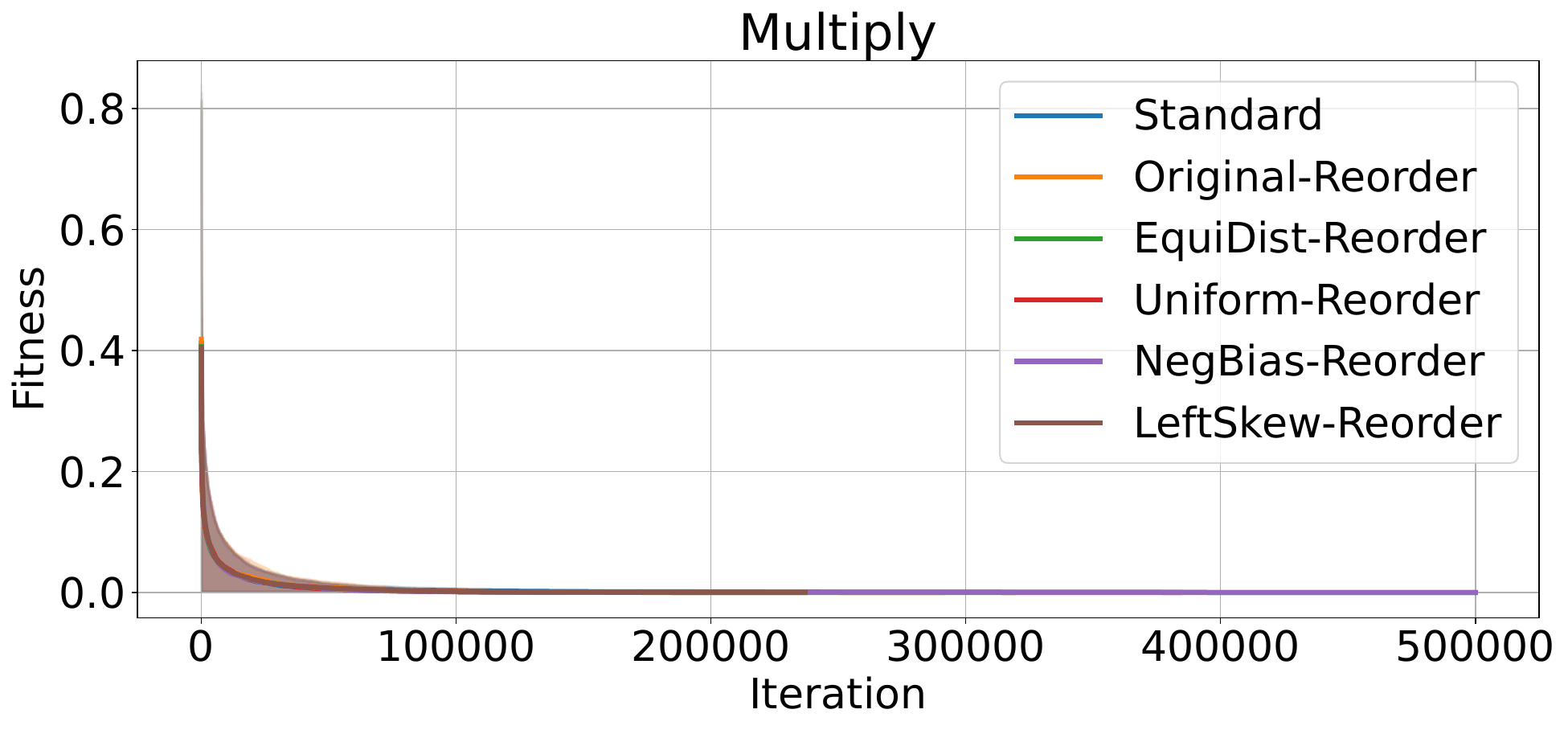}
        
        \caption{Convergence plots for each symbolic regression and Boolean benchmark. The shaded area indicates their respective standard deviation. The x- and y-axis are scaled to make the standard deviation visible. }
        \label{fig_appendix:convergence_plots_with_std}
    \end{figure}

    For further information and visual inspection, convergence plots without changes to the y- and x-axis are provided in Figure~\ref{fig_appendix:convergence_plots}.
    As is already mentioned in Section~\ref{subsec:conv_behaviour}, their convergence behaviour can be classified as \emph{Fast to Slow}~\cite{Stegherr23}.
    This categorization is easier to see when the axis are not scaled. 
    
    Furthermore, Figure~\ref{fig_appendix:convergence_plots_with_std} include their respective standard deviation.
    We included them as separate plots to increase visibility.
    Please note that the x- and y-axis are logarithmically scaled again.
    Otherwise, the standard deviation would not be visible.
    For both benchmarks, the standard deviation is relatively low.
    In the context of symbolic regression benchmarks, the CGP variants have a higher standard deviation at the beginning and at the end of its training.
    Boolean benchmarks only show a higher standard deviation during the beginning of the training, and little to no deviation during the rest.

\end{appendices}



\end{document}